\documentclass[]{jdmdh}

\usepackage{array}
\usepackage{pgfplots}
\pgfplotsset{compat=1.16}
\usepackage{lscape}
\usepackage{tikz}
\usetikzlibrary{arrows.meta}
\tikzset{>={Latex[width=3mm,length=3mm]}}
\usetikzlibrary{shapes,positioning}

\usepackage{longtable}
\usepackage{microtype}

\title{Corpus and Models for Lemmatisation and POS-tagging\\ of Classical French Theatre}

\author[1]{Jean-Baptiste Camps}
\author[2]{Simon Gabay}
\author[3]{Paul Fièvre}
\author[1]{Thibault Clérice}
\author[4]{Florian Cafiero}

\affil[1]{Centre Jean-Mabillon, École nationale des chartes, Université Paris, Sciences \& Lettres}
\affil[2]{Université de Neuchâtel}
\affil[3]{Bibliothèque Nationale de France}
\affil[4]{GEMASS, CNRS / Université Paris-Sorbonne}

\corrauthor{Jean-Baptiste Camps}{Jean-Baptiste.Camps@chartes.psl.eu}

\begin{document}
\maketitle
\abstract{This paper describes the process of building an annotated corpus and training models for classical French literature, with a focus on theatre, and particularly comedies in verse. It was originally developed as a preliminary step to the stylometric analyses presented in \citet{Cafieroeaax5489}. The use of a recent lemmatiser based on neural networks and a CRF tagger allows to achieve accuracies beyond the current state-of-the art on the in-domain test, and proves to be robust during out-of-domain tests, \textit{i.e.} up to 20th\,c. novels.}

\section{Introduction}
\label{sec:introduction}
\strut

If many lemmatisers and POS taggers have been trained, and sometimes conceived, for  French (\textit{e.g.} \cite{tellier2012segmenteur}, \cite{urieli2013}\dots), they usually focus on contemporary French and tools for \textit{Ancien Régime} French remain scarce. One important exception is the \textit{TreeTagger} \citep{Schmid1995} model developed by \cite{diwersy_ressources_2017} for the \textit{Presto} project \citep{presto}. They have prepared training data with c.\,60,000 tokens annotated manually, using an adapted version of \textsc{Multex} \citep{Ide1994} and \textsc{Grace} \citep{Adda1998} for the parts of speech (POS), and the Le\textit{fff} \citep{Sagot2010}, \textit{Morphalou} \citep{Romary2004} and \textit{LGeRM} \citep{Souvay2009} for the lemmas. Unfortunately, training data are not publicly available (yet?)\footnote{%
The \textit{Presto} data have been entirely revised after the initial publication of this paper (cf.~\cite{gabay_standardizing_2020}) and are available online at the following address: \url{https://github.com/e-ditiones/LEM17}. %
}, they are made of non-normalised texts from the 16th to the 18th\,c. which prevents any comparison task, and no detailed evaluation of the model is offered by the designers of the model.

Recent developments have renewed the field of normalisation, lemmatisation and tagging of historic language varieties, including variation rich languages. Particularly, deep learning methods have been used to establish a new state of the art for Middle Dutch \citep{kestemont_lemmatization_2016}, 
Medieval Latin \citep{kestemont_integrated_2017}, Medieval Occitan \citep{camps_production_2017}, Early Irish \citep{dereza2019lemmatisation} or Middle High German and a variety of other languages \citep{schmid2019deep,manjavacas_improving_2019}.
Convolutional or (bidirectional) LSTM networks have been put to use, in architecture involving word or character levels representations, for instance word embeddings, and usually a sequence to sequence approach using an encoder–decoder model \citep{bollmann_large-scale_2019}, that allow for taking into account the context of a token at either word or character level. 

Such architectures are now competing with character level statistical machine translation models for the state of the art of historical text normalisations \citep{tang_evaluation_2018,bollmann_large-scale_2019,gabay:hal-02596669}.
The developments of such architectures allowed to
obtain results in the 89 to 96\% range for lemmatisation of historical languages (and even up to 97.86\% for standard languages like modern French, see \cite{manjavacas_improving_2019}), though they 
 have been showed to be the most dependent on the quantity and quality of training data, outperforming more traditional (essentially, rule based) approaches only when substantial training data is available \citep{bollmann_large-scale_2019}.

POS and morphological taggers, especially in the case of morphologically rich languages, have also benefited from neural architectures, with state of the art for modern languages often being attained by recurrent neural nets \citep{ling_finding_2015,heigold_neural_2016,schmid2019deep},
 with accuracies in the range of 97-98\% for POS and 92-94\% for morphology \citep{heigold_neural_2016}, and more specifically from 90 to 98\% for POS tagging of historical languages  \citep{schmid2019deep,manjavacas_improving_2019}.

In this paper, we present in detail our corpus, the annotation choices, the training set up and finally the results obtained by two models. On the one hand, an ``extended'' model for lemmatisation of normalised 17th\,c. French theatre, \textit{i.e.} tested specifically for this data, but with enough additional training material to have it perform relatively well on modern (16th\,c. to 18th\,c.) and even contemporary (19th\,c. to 20th\,c.) French. On the other hand a POS-tagging model including the morphology (gender, number, tense\dots), based exclusively on 17th\,c. century training data.

These models were initially developed to allow the stylometric analyses presented in \cite{Cafieroeaax5489},  especially the analyses based on lemma, lemma in rhyme position, and POS 3-grams. In this context, getting results as accurate as possible was a necessary step to ensure the reliability of stylometric results. This initial motivation is reflected in some of the choices of this paper, notably the constitution of the training corpus (which is a sample from the corpus that was to be used in the analyses)\footnote{%
Stylometric results are not presented in this paper. For them, see \cite{Cafieroeaax5489}.%
}.

\section{The \textit{Théâtre Classique} Corpus}
\strut

Because of its cultural importance for French literature and thanks to a few innovative pioneers, 17th\,c. French theatre (also called \textit{théâtre classique}, ``classical theatre'') benefits from many digital editions freely available online. 
Among the various projects offering texts online, the \textit{Théâtre classique} database \citep{fievre_theatre_2007,schoch-2018} proposes one of the most comprehensive collections. The texts derive from digitisations of 17th\,c. prints taken from \textit{Gallica} \citep{gallica}. The oldest print is usually used, but it is not necessarily the case (for detailed bibliographic information on 17th\,c. printed theatre, cf.\,\citet{riffaud2014}). The text is normalised manually and the spelling aligned with contemporary French\footnote{About this (strange) French tradition, cf.~\cite{gabay_pourquoi_2014,duval_les_2015}.}

Among all the plays available, 41 comedies (cf.\,appendix~\ref{annexe:corpusMain}) have been selected to create a training corpus. They have been written by six different authors spread over two generation:
Antoine Le Métel d’Ouville (1589-1655), Jean de Rotrou (1609-1650), Paul Scarron (1610-1660), Pierre Corneille (1606-1684), Molière (1622-1673) and Thomas Corneille (1625-1709). All the plays have been written between the 1630's and the 1670's, that is to say within c.\,40 years.

\section{Building an annotated corpus}
\strut

The annotation scheme has been conceived to cope with diachronically spread data, especially earlier periods such as middle and Renaissance French.

\subsection{Choice of authority lists}
Since annotation principles are rather complex, we have decided to publish guidelines separately~\citep{gabay2020}, and we will summarise here our most important choices.

Regarding POS tags, on top of the aforementioned \textsc{Multex}, many possibilities exist: \textsc{Eagles} \citep{Leech1996}, \textsc{Ud-Pos} \citep{Petrov2011} and \textsc{Cattex} \citep{prevost_jeu_2013}. While EAGLES and UD-POS have been developed as international standards, \textsc{Cattex} has been designed specifically for French medieval texts.

We have decided to choose \textsc{Cattex}, because we are interested in a long diachronic perspective, and we want to maintain interoperability with several existing corpora that are already using it, such as the \textit{Base de français médiéval} (``Medieval French Database'', cf.~\citet{Guillot2017}) or the \textit{Geste} corpus \citep{camps_geste:_2019}\footnote{%
It is to be noted that using Cattex does not prevent, in the future, conversions to other tag-sets. For instance, the SRCMF corpus \citep{prevost_syntactic_2013}, originally tagged with Cattex tags, exists in a converted version using UD-POS \citep{noauthor_universaldependenciesud_old_french-srcmf_2020}.
}.

The annotation manual of \textsc{Cattex09} (\citet{prevost_principes_2013}) offers a detailed list of tagging rules that we strictly observed. Three options are offered to the annotator: morphological tagging, morpho-syntactical tagging\footnote{With morpho-syntactical tagging, categories are mainly determined in context: \textit{il veut le \textbf{bien}} (\textit{bien} is \texttt{NOMcom}) vs \textit{il est \textbf{bien} grand} (\textit{bien} is \texttt{ADVgen}).}, or both. If adding both labels is ideal (to study processes such as adjectivisation, substantivisation\dots) it remains far to costly in time, and we have therefore opted for a simple morpho-syntactical tagging, that appeared at the time as an interesting middle way.

Regarding lemmatisation, we have already mentioned \textit{LGeRM}, the Le\textit{fff} and \textit{Morphalou}. The main interest of the last one, that we have chosen, is that not only the v3.1 includes the Le\textit{fff}, but it is also used by the \textit{Frantext} base – the data of which is partially available online to be used as additional material for our model – and the \textit{Trésor de la Langue Française informatisé}. The \textit{LGeRM} lexicon is irrelevant in our case, since it is an artificially archaised version of \textit{Morphalou} to match 17th and 18th\,c. forms. Concerning proper names, we built a specific reference list, thanks to the characters and places index provided by \cite{fievre_theatre_2007}, that we expanded when necessary.

Some of our choices diverge from those made by the authors of \textit{Morphalou}. We were, for instance, more systematic in choosing the masculine singular form as a lemma for nouns (\textit{baronne} is lemmatised as \textit{baron}) but not only (\textit{la} (det. def.) as \textit{le}, \textit{sa} or \textit{ses} (poss.) as \textit{son}). 
Concerning personal pronouns, the singular masculine subject (when relevant) case has been used as lemma: direct regimen forms (\textit{le}, \textit{la}, \textit{les}, as in, \textit{il les donne}) as well as indirect regimen forms (\textit{lui}, \textit{elle(s)}) have been lemmatised to the subject masculine singular (\textit{il}) – one single pronoun can indeed be subject (\textit{il}), reflexive (\textit{se}), direct object (\textit{le} or \textit{en}), indirect object (\textit{lui} or \textit{y}) or disjunctive (\textit{lui}). Still in line with our diachronic approach, we kept the difference between the old partitive \textit{des} (contracted form of \textit{de les}) and the new non-definite plural article \textit{des}, and encoded the contracted form \textit{au(x)} as \textit{a+le}.

It has to be noted that, since lemmas are not numbered in \textit{Morphalou}, it has not been possible to introduce a number-based disambiguation for homographs (e.g., \textit{son1} (poss. ``his'') vs. \textit{son2} (noun ``sound'')\dots). Such a situation is however only partially problematic, since it remains possible to distinguish forms thanks to the POS, or the morphology.

\subsection{Texts preprocessing and sampling}

In order to limit model biases, each play of the corpus was sampled to create training and testing data. The text of \cite{fievre_theatre_2007} editions have been tokenised using \textit{TXM} \citep{heiden_txm_2010} \textsc{xml} import. During the import an XSL filter was used to retain only the character's speeches, with exclusion of all other material (stage directions, act and scene numbering\dots).
Out of these data, a three-tier sample was constituted with the 2,000 first tokens of our 41 plays for training data (hereafter train set), the 100 median tokens for validation data (hereafter dev set), and the last 100 for testing data (hereafter test set). 
Case was not normalised, in order to keep information relevant to the identification of proper names. 

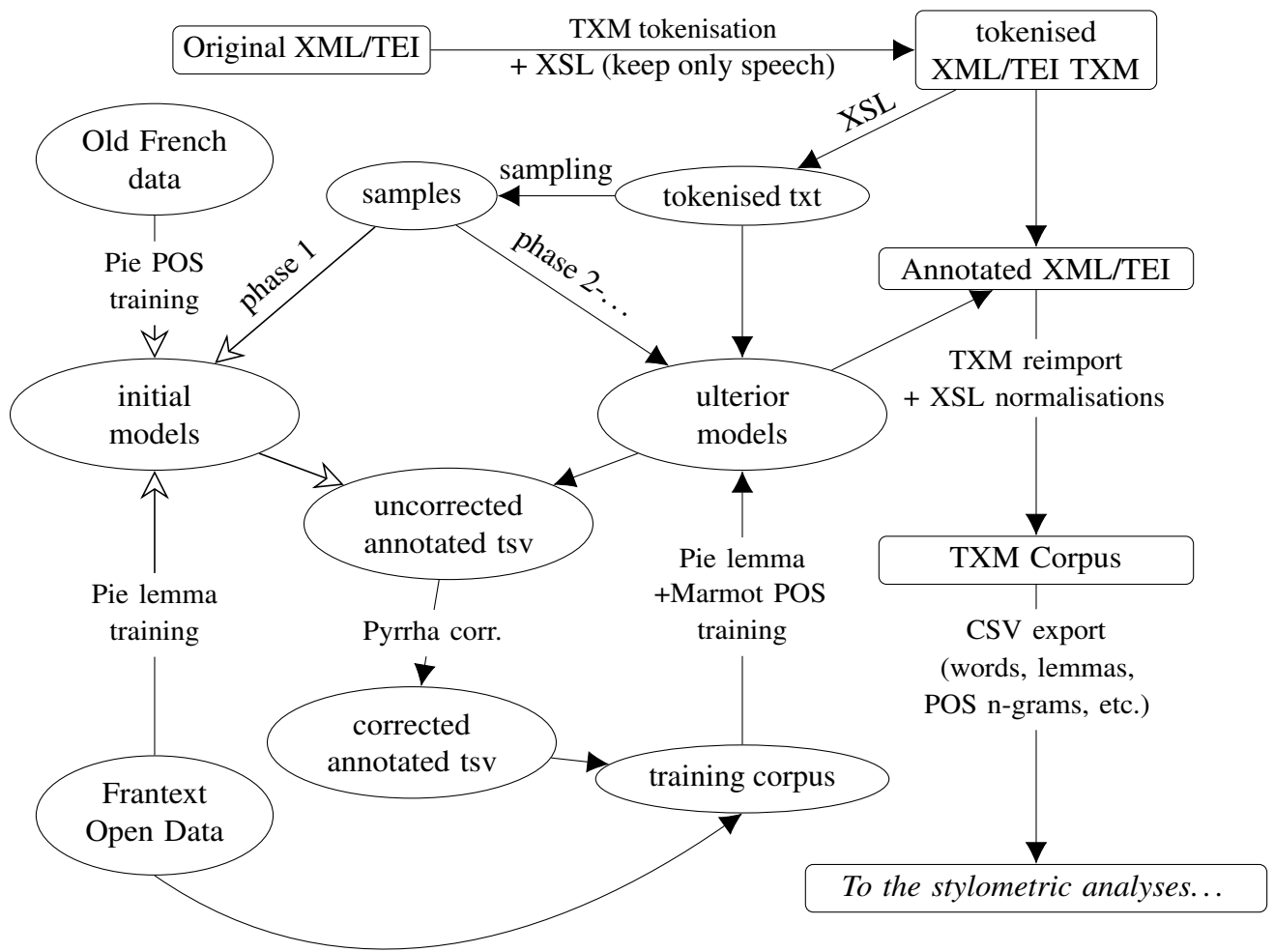
\begin{figure}[htbp]
    \centering
\begin{tikzpicture}
\node[draw, rectangle, rounded corners=3pt, text centered] (0) at (-10, 10.5)
{Original XML/TEI};
\node[draw, rectangle, rounded corners=3pt, text width=3cm, text centered] (A) at (0, 10.5) {tokenised XML/TEI TXM}; 
\node[ellipse, draw, text centered] (b) at (-4, 8.5) {tokenised txt}; 
\node[ellipse, draw, text centered] (c) at (-8.5, 8.5) {samples}; 
\node[ellipse, draw, text centered, text width=2cm] (alpha) at (-12, 9) {Old French data}; 
\node[text centered, text width=2cm] (alphaprime) at (-12, 7.3) {\small Pie POS training};
\node[draw, ellipse, align=center, text width=2cm] (gamma) at (-12, 0)
{Frantext Open Data};
\node[align=center, text width=2cm] (gammaprime) at (-12, 2.75)
{\small Pie lemma training};
\node[ellipse, draw, text width=2.5cm, text centered] (d) at (-8, 4) {uncorrected annotated tsv};
\node[text width=2.5cm, text centered] (dprime) at (-8.25, 2.5) {\small Pyrrha corr.};
\node[ellipse, draw, text width=2.5cm, text centered] (e) at (-8.5, 1) {corrected annotated tsv};
\node[ellipse, draw, text centered] (f) at (-4, 0.5) {training corpus};
\node[text centered, text width=2.5cm] (fprime) at (-4, 3) {\small Pie lemma\\+Marmot POS\\training};
\node[ellipse, draw, text centered, text width=2.5cm] (beta) at (-12, 5.5) {initial\\ models};
\node[ellipse, draw, text width=2.5cm, text centered] (g) at (-4, 5.5) {ulterior\\ models};
\node[draw, rectangle, rounded corners=3pt, text width=4cm, text centered] (B) at (0, 7.5) {Annotated XML/TEI};
\node[text width=6cm, text centered] (i) at (0, 6) {\small TXM reimport\\ + XSL normalisations};
\node[draw, rectangle, rounded corners=3pt, text width=4cm, text centered] (C) at (0, 3.5) {TXM Corpus};
\node[text width=4cm, text centered] (ii) at (0, 2) {\small CSV export\\(words, lemmas, POS n-grams, etc.)};
\node[draw, rectangle, rounded corners=3pt, text width=6cm, text centered] (D) at (0, -1) {\itshape To the stylometric analyses…};

\draw[->] (0) -- (A) node[midway, , align=center] {\small TXM tokenisation\\ + XSL (keep only speech)};
\draw[->] (A) -- (B);
\draw[-] (B) -- (i);
\draw[->] (i) -- (C);
\draw[-,] (C) -- (ii);
\draw[->] (ii) -- (D);
\draw[->] (A) -- (b) node[midway,above,sloped] {XSL};
\draw[->] (b) -- (c) node[midway,above,sloped] {sampling};
\draw[-{Stealth[scale=1.3,angle'=45,open,width=3mm,length=3mm]},semithick] (c) -- (beta) node[midway,above,sloped] {phase 1};
\draw[-{Stealth[scale=1.3,angle'=45,open,width=3mm,length=3mm]},semithick] (beta) -- (d);
\draw[-] (d) -- (dprime);
\draw[->] (dprime) -- (e);
\draw[->] (e) -- (f) 
;
\draw[-] (f) -- (fprime);
\draw[->] (fprime) -- (g);
\draw[->] (c) -- (g) node[midway,above,sloped] {phase 2-…};
\draw[->] (b) -- (g);
\draw[->] (g) -- (d);
\draw[->] (g) -- (B);
\draw[-] (alpha) -- (alphaprime);
\draw[-{Stealth[scale=1.3,angle'=45,open,width=3mm,length=3mm]},semithick] (alphaprime) -- (beta);
\draw[-] (gamma) -- (gammaprime);
\draw[-{Stealth[scale=1.3,angle'=45,open,width=3mm,length=3mm]},semithick] (gammaprime) -- (beta);
\draw[->] (gamma.south) to [out=-35,in=-145] (f.south);

\end{tikzpicture}
    \caption{Workflow for building the training corpus and models (ellipses) and the annotated XML corpus used for stylometric analysis (rectangles); void-arrowheads specify steps particular to the initial model building phase}
    \label{fig:workflow}
\end{figure}

The complete XML and annotation workflow is presented in fig.~\ref{fig:workflow}\footnote{%
   Since the initial publication of this article, the effects of Unicode NFKD normalisation have also been tested, with an unclear effect on training accuracy \cite{gabay_standardizing_2020}.
}.

\subsection{Annotation and correction process}

The annotation has been done in three phases.
A first \textit{Pie} \citep{pie} lemmatisation model has been trained only on the \textit{Frantext Open Access} data \citep{frantext}, and has been used to annotate a first sample of c.\,40,000 tokens, in combination with an available \textit{Pie} model for POS tags trained on Old French\footnote{%
    A recent version of the model for Old French can be found as part of the web application \textit{Deucalion} \citep{deucalionAF}; they are also directly usable through \textit{Pyrrha}'s interface \citep{thibault_clerice_2019_2541730}.
    Finally, the most up-to-date version of both the Old French and Classical French models is provided, along with functionalities to ease tagging of new documents, as part of the simple \texttt{pie-extended} Python Package \citep{pie-extended}. The models can be procured using their linguistic code (\texttt{fro} for Old French, \texttt{fr} for Classical French) by running the commands:
        \texttt{pie-extended download fr}
    and texts can be annotated by
        \texttt{pie-extended tag fr MyFile.txt}.
        They also are available online, on the École des chartes' \textit{Deucalion} instance, at \url{https://dh.chartes.psl.eu/deucalion/}.
}.
After being corrected, the same corpus has been used to train new models and annotate c.\,40,000 other tokens that were, once again, corrected to create the final training corpus.

\begin{figure}[htbp]
  \centering
  \includegraphics[width=\textwidth,height=0.5\textheight,keepaspectratio]{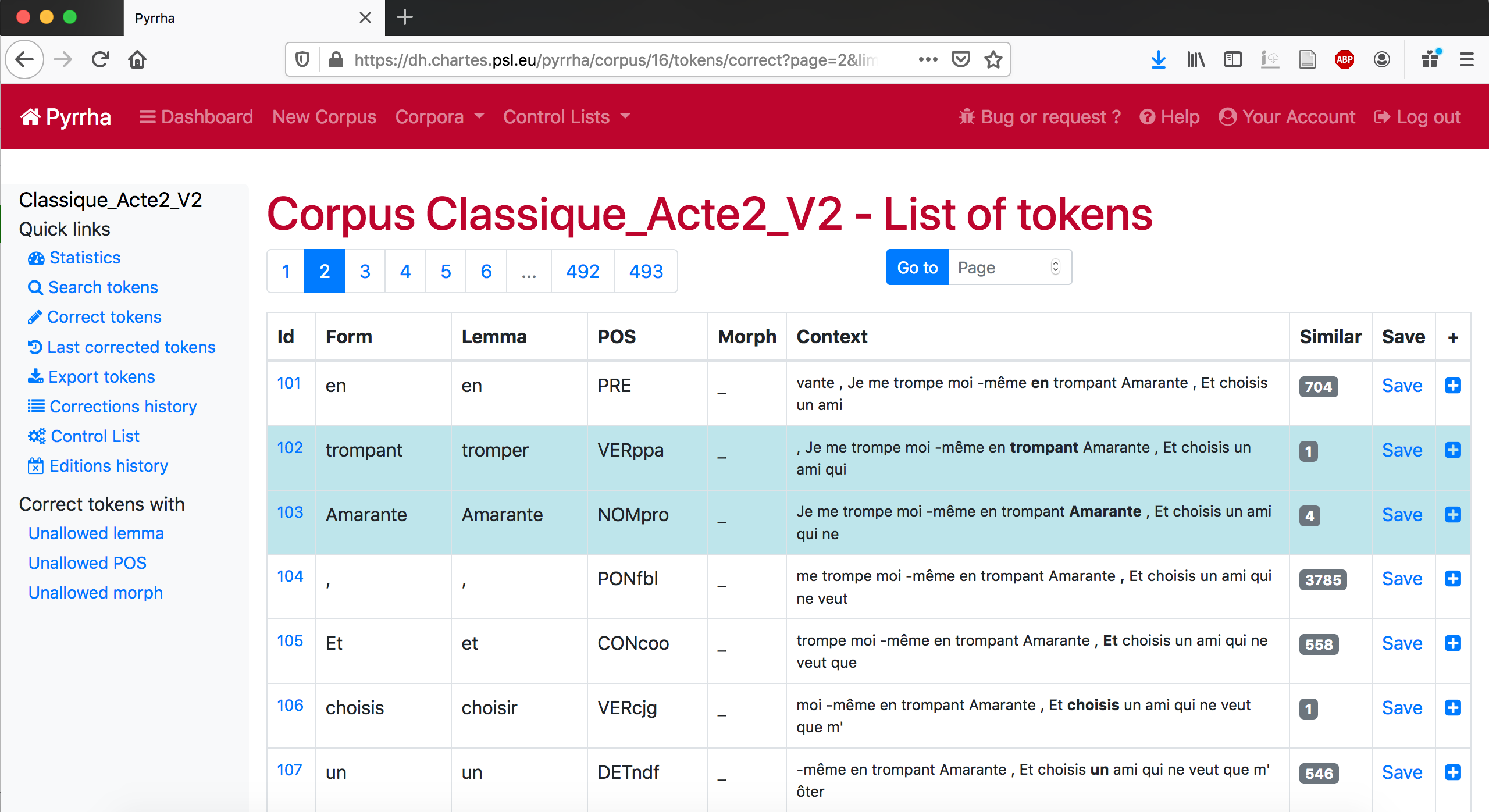}

  \caption{Screenshot of the \textit{Pyrrha} interface: main correction view}
  \label{fig:pyrrha}
\end{figure}

\begin{figure}[htbp]
  \centering
  \includegraphics[width=\textwidth]{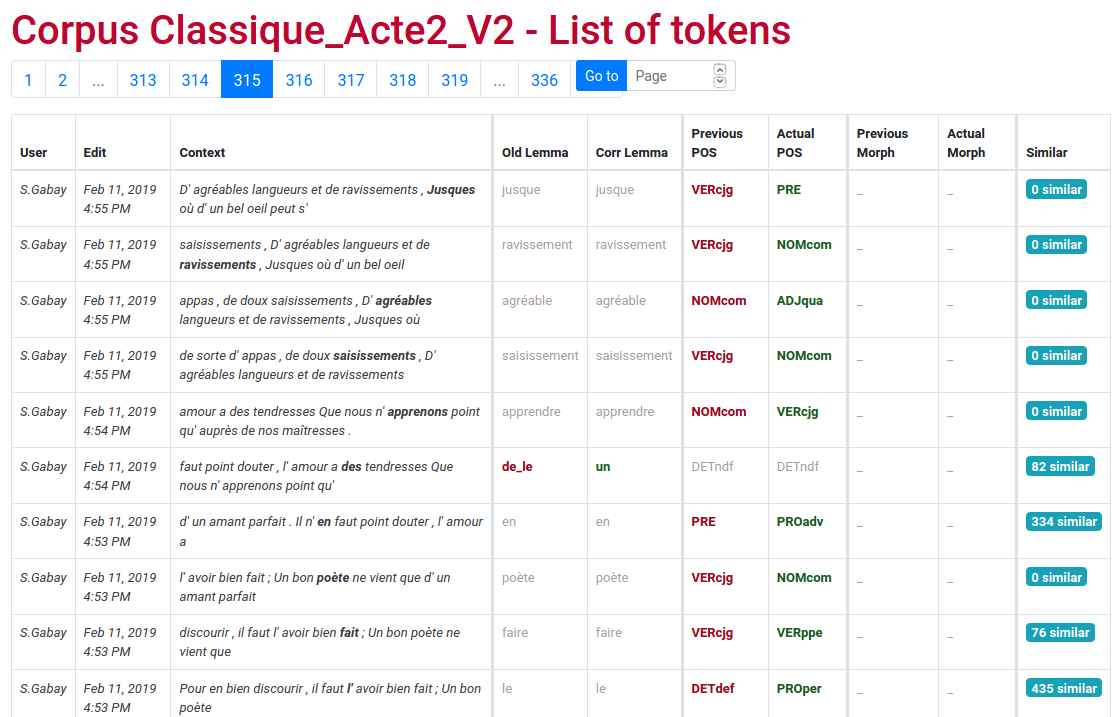}
  
  \smallskip
  
  \hspace*{-1cm}\includegraphics[width=1.2\textwidth]{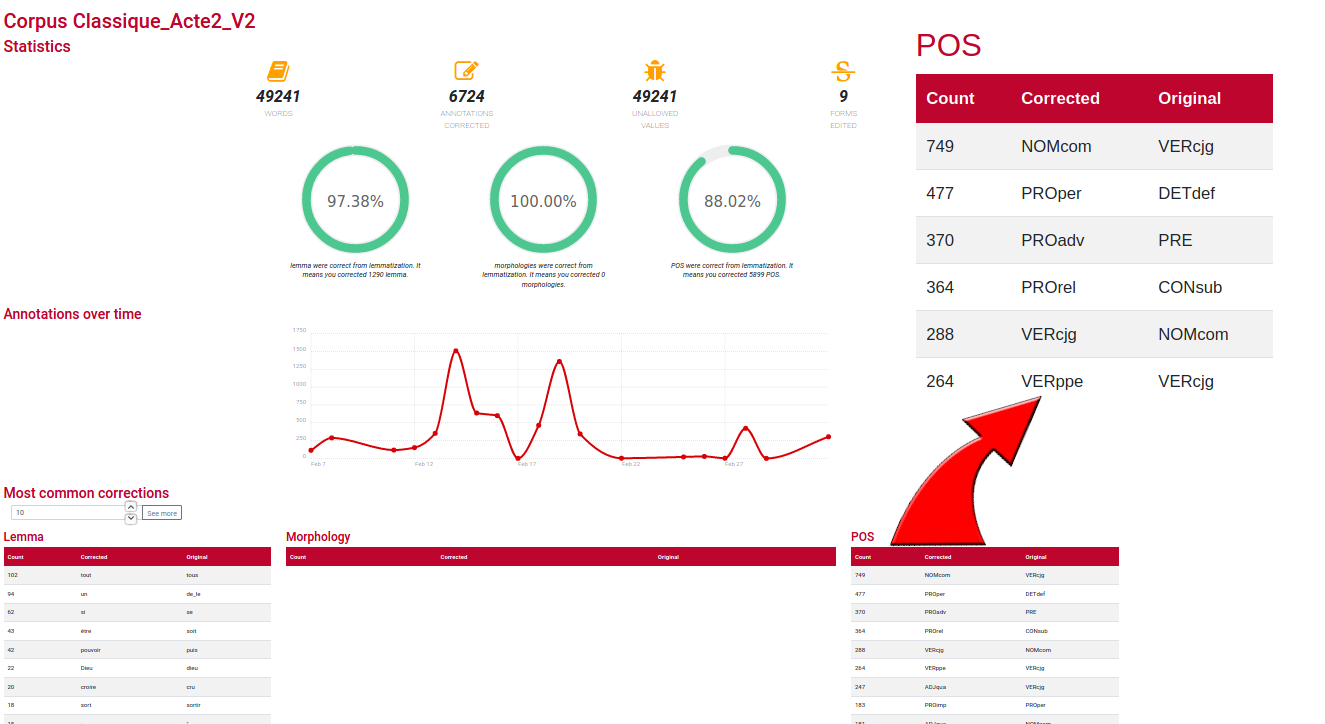}
  \caption{Screenshot of the \textit{Pyrrha} interface: correction history and corpus statistics}
  \label{fig:pyrrha_corrs}
\end{figure}

Lemma and POS-tags have all been corrected manually. This work was facilitated by the use of \textit{Pyrrha} \citep{thibault_clerice_2019_2541730}, a web-based correction interface able to do batch correcting as well as to handle authority lists, allowing efficient collaborative work (fig.~\ref{fig:pyrrha}). \textit{Pyrrha} also keeps tracks of all changes made on the corpus (fig. \ref{fig:pyrrha_corrs}), and makes it possible to import, correct, share, and download back corpora and authority lists. 

\subsection{Expanding annotation through available resources}

If POS-tags have all been systematically corrected, through both linear reading and batch corrections, it is not the case of the morphology, which has only been mostly batch-corrected, because of time concerns.
Thus, we can guarantee that every POS tag has been proofread at least once (and usually multiple times), which is not the case for the morphology.

Indeed, to save time, morphological information was not added manually, but was instead projected using the lexicon of inflected forms \textit{Morphalou} \citep{morphalou}. To do so, \textsc{Cattex} POS-tags were mapped to \textit{Morphalou}'s categories (table \ref{tab:mapIn})\footnote{%
    Punctuation and proper names were dealt with separately and specifically, as well as \textsc{Cattex} combined label, \texttt{PRE.DETdef}.
    The correspondence between some categories reflect substantial difference in the grammatical typology used. For instance, \textsc{Cattex} \texttt{PROord} (e.g., `premier' in `le premier') match Morphalou's more generic common noun category.%
    For more information on \textsc{Cattex} categories, see \cite{guillot_manuel_2013}.
}. Then, a simple algorithm looked for matching forms inside \textit{Morphalou}: when the form was unambiguous, the morphological information was directly retrieved, otherwise the hand-corrected POS was used to assess the correct morphological information to retrieve. If none was found, an \texttt{unknown} morph tag was added.

\begin{table}[htbp]
    \centering \footnotesize %
    \begin{tabular}[t]{rr|}
\textbf{\textsc{Cattex}} & \textbf{\textit{Morphalou}} \\ \hline \hline
INJ&Interjection\\ 
ADVgen&Adverbe\\ 
ADVneg&Adverbe\\ 
ADVint&Adverbe\\ 
ADVsub&Adverbe\\ 
CONcoo&Conjonction\\ 
CONsub&Conjonction\\ 
VERcjg&Verbe\\ 
VERinf&Verbe\\ 
VERppe&Verbe\\ 
VERppa&Verbe\\ 
PRE&Préposition
    \end{tabular}
\begin{tabular}[t]{|cc|}
    \textbf{\textsc{Cattex}} & \textbf{\textit{Morphalou}} \\ \hline \hline
DETdef&Déterminant\\ 
DETndf&Déterminant\\ 
DETdem&Déterminant\\ 
DETpos&Déterminant\\ 
DETind&Déterminant\\ 
DETrel&Déterminant\\ 
DETint&Déterminant\\ 
DETcom&Déterminant\\ 
PROper&Pronom\\ 
PROimp&Pronom\\ 
PROadv&Pronom\\ 
PROpos&Pronom\\ 
PROdem&Pronom\\ 
PROind&Pronom
    \end{tabular}
\begin{tabular}[t]{|ll}
\textbf{\textsc{Cattex}} & \textbf{\textit{Morphalou}} \\ \hline \hline
PROord&Nom commun\\ 
PROrel&Pronom\\ 
PROint&Pronom\\ 
PROcom&Déterminant\\ 
ADJqua&Adjectif qualificatif\\ 
ADJind&Adjectif qualificatif\\ 
ADJord&Adjectif qualificatif\\ 
ADJpos&Adjectif qualificatif\\ 
NOMcom&Nom commun\\ 
ADJcar&Nombre\\ 
DETcar&Nombre\\ 
PROcar&Nombre
\end{tabular}
    \caption{Mapping of \textsc{Cattex} POS-tags to \textit{Morphalou} categories}
    \label{tab:mapIn}
\end{table}

Finally, Morphalou's information was converted back to \textsc{Cattex09} format, using the mapping presented in table \ref{tab:mapOut}.

\begin{table}[htbp]
    \centering \footnotesize %
\begin{tabular}[t]{cc}
\textbf{\textit{Morphalou}} & \textbf{\textsc{Cattex}} \\ \hline \hline
\multicolumn{2}{c}{\textbf{Mode}}\\ \hline 
indicative&ind\\ 
imperative&imp\\ 
conditional&con\\ 
subjunctive&sub\\ 
infinitive&-\\ 
past&-\\ 
participle&-
    \end{tabular}
\begin{tabular}[t]{cc}
\textbf{\textit{Morphalou}} & \textbf{\textsc{Cattex}} \\ \hline \hline
\multicolumn{2}{c}{\textbf{Temps}}\\ \hline
present&pst\\ 
imperfect&ipf\\ 
future&fut\\ 
simplePast&psp\\ \hline
\multicolumn{2}{c}{\textbf{Pers.}}\\ \hline
firstPerson&1\\ 
secondPerson&2\\ 
thirdPerson&3
\end{tabular}
\begin{tabular}[t]{cc}
\textbf{\textit{Morphalou}} & \textbf{\textsc{Cattex}} \\ \hline \hline
\multicolumn{2}{c}{\textbf{Nomb.}}\\ \hline
singular&s\\ 
plural&p\\ \hline
\multicolumn{2}{c}{\textbf{Genre}}\\ \hline
masculine&m\\ 
feminine&f\\ 
neuter&n\\ \hline
\multicolumn{2}{c}{\textit{Varia}}\\ \hline
-&-\\ 
invariable&x\\ 
1036442&ERROR
\end{tabular}
    \caption{Mapping of \textit{Morphalou} to \textsc{Cattex} flexion tags.}
    \label{tab:mapOut}
\end{table}

\subsection{Increasing corpus generality with \textit{Frantext} data}
The \textit{Frantext} base offers an open access version \citep{frantext}, 32 texts of which have been used to increase the training data (see appendix~ \ref{annexe:frantext_openaccess}).
Marginal interventions have been made to correct some systematic errors, but also to ensure its consistency with our annotation choices. For instance, for pronouns (labelled as \texttt{CLO}, \texttt{CLS} and \texttt{PRO} in \textit{Frantext} data), the lemmas \textit{je}, \textit{me}, \textit{m'}, \textit{M'}, \textit{moi}, \textit{Moi}, were mapped to \textit{je}; likewise, \textit{ils}, \textit{elle}, \textit{elles}, \textit{le}, \textit{la}, \textit{les}, \textit{lui}, \textit{leur}, \textit{eux}, \textit{Ils}, \textit{Lui}, \textit{Elle}, \textit{Elles}, to \textit{il}, etc. On the other hand, some forms of \textit{celui} and \textit{cela} were originally lemmatised to \textit{il}, and we changed the lemmatisation to \textit{celui} and \textit{cela}.
Similarly, forms of \textit{chacun} (or \textit{aucun}) were lemmatised to \textit{un}, and we changed it to \textit{chacun} (or \textit{aucun}).

We were also more systematic in the alignment of feminine and masculine forms to a single (masculine, singular) lemma, may it be for possessives (\textit{mienne}, \textit{tienne}, \textit{sienne} to \textit{mien}, \textit{tien}, \textit{sien}) or just nouns (\textit{hôtesse}, \textit{amie}, \textit{veuve}, \textit{captive}, to \textit{hôte}, \textit{ami}, \textit{veuf}, \textit{captif}, etc).

A few minor corrections of obvious errors were also made (\textit{e.g.}, \textit{saurer} to \textit{savoir}), especially regarding homograph forms of some lemmas (\textit{e.g.}, between verbal forms of \textit{défaire}, ``undo',' and the noun \textit{défaite}, ``defeat'', or between \textit{ver}, ``worm'' and \textit{vers}, ``verse''). An additional adjustment has also been made regarding the ligature \textit{œ} (\textit{cœur}), which has been preferred over its decomposed form (\textit{coeur}).

\section{Training setup}
\label{sec:training}
\strut

As previously mentioned, many tools are already available. \textit{TreeTagger} \citep{Schmid1995} remains one of the most widely used, even though it is outperformed by other solutions. For the French language, \textit{TALISMAN}~\citep{urieli2013} or \textit{MElt}~\citet{sagot2016} are commonly used by the NLP community, but many other solutions are available such as \textit{Lemming} (only for lemmas, cf.~\citet{muller-etal-2015-joint}), \textit{Marmot} (only for POS, cf.~\citet{muller2013efficient}) or \textit{Pie} (mainly for lemmas, cf.~\citet{pie}). We have decided to use the two latter.

\subsection{Lemmatisation}
\label{sec:training-lemmatisation}

Concerning \textit{Pie} as a lemmatiser, we tested three different configurations (table~\ref{tab:configs_pie}):

\begin{enumerate}
    \item \textbf{base (sent-lm)}, default configuration, based on the configuration that achieved best accuracy described in \cite{pie}, using sentence context, RNN character embeddings, as well as forward and backward language models\footnote{%
    For this configuration, we used a config file provided to us by Enrique Manjavacas, main developer of \textit{Pie}.
    }; 
    \item \textbf{wembs} the same as the previous one, but with the adjunction of 
    word embeddings, initialised using pretrained embeddings;
    \item \textbf{bert} same as the previous one, but using \textit{CamemBERT} embeddings \citep{camembert}, reduced from 768 to 150 dims; 
    \item \textbf{cnn+wembs} a configuration using CNN character embeddings, with word embeddings, based on \texttt{skipgram} on a larger unlemmatised corpus. This configuration is the one used for \cite{Cafieroeaax5489}, with limited additional tuning. 
\end{enumerate}

\begin{table}[htbp]
    \centering \small %
    \begin{tabular}{c|cccc}
         & \textbf{context} & \textbf{char embs} & \textbf{word embs} & \textbf{hidden size} \\ \hline \hline
    base & sentence & rnn 300 dims  & 0 & 150 \\
    wembs & sent.+word & rnn 300 dims & 150 dims  & 150\\
    bert & sent.+word & rnn 300 dims & 768 to 150 dims  & 150\\
    cnn+wembs & sent.+word & cnn 150 dims & 150 dims & 150
    \end{tabular}
    \caption{Configurations tested with \textit{Pie}}
    \label{tab:configs_pie}
\end{table}

For each configuration, due to the stochastic nature of the process, five models were trained, using early stopping with threshold 0.001 and patience 6, and the best one was retained.

For the third configuration, word-embeddings were pretrained using \textit{Word2Vec} Python implementation, 
on a large corpus of 343 theatre texts from \cite{fievre_theatre_2007} and those of the \textit{Frantext Open Access} that we presented \textit{supra}, for a total of c.\,7M tokens.

\clearpage

\subsection{POS tagging}

For POS-tagging, we trained both \textit{Marmot} and \textit{Pie} on the training data we produced, without further augmentation. The configurations were the following:

\begin{enumerate}
    \item \textbf{Marmot}: base configuration provided with \textit{Marmot}, using the dev set during training, and the test set for final evaluation;
    \item \textbf{Pie}: same configurations that for lemmatisation (\textbf{base (sent-lm)},\textbf{wembs} and \textbf{bert}, see above and table~\ref{tab:configs_pie}) but with a CRF output layer to predict part-of-speech tags;
    \item \textbf{+aux}: for each of the POS-tagging configuration, we tried to see if we could obtain a gain in accuracy by using auxiliary tasks (the tasks used can be seen in table \ref{tab:mapOut} in bold, with the addition of \textbf{Case}, mainly used for personal pronouns). 
    In a  multi-task setting, we trained linear classifiers for each morphological feature, but sharing weights with the main task (POS prediction).
\end{enumerate}

\section{Results}
\strut

\subsection{Calibration and in-domain tests}

Results of the \textit{Pie} lemmatisation training are presented in table~\ref{tab:pie_accs}. The best \textit{Pie} model (configurations 2, \textbf{wembs}) achieved 99.09\% accuracy on the test set. The results obtained with \textit{Marmot} and \textit{Pie} for POS on the test set are presented in table~\ref{tab:marmot-results}, and the best results are achieved by the \textbf{wembs+aux} configuration. Yet, in both cases, the variation between the accuracies are relatively low, and not substantially higher than random variations between different iterations of the same configuration.

\begin{table}[htbp]
    \centering \small %
    \begin{tabular}{c|rrrrr}
& \textbf{base} & \textbf{wembs} & \textbf{bert} & \textbf{cnn-wembs} & \textit{support} \\ \hline \hline
\textit{all} & 98.80 & \textbf{99.09} & 98.95 & 98.80 &  \textit{4181}\\
\textit{unknown tokens} & 70.31 & 71.88 & \textbf{76.56} & 65.62 & \textit{64} \\
\textit{ambiguous tokens} & 97.32 & \textbf{98.02} & 97.43 & 97.43 & \textit{857} \\
\textit{unknown targets} & 50.00 & 57.14 & \textbf{85.71} & 57.14 & \textit{14}
    \end{tabular}
    \caption{\textit{Pie} lemmatisation accuracies on the test set for the best model for each configuration. ``Unknown tokens'' are tokens never seen during training, while ``ambiguous tokens'' are forms that can correspond to different lemmas. ``Unknown targets'' are lemmas never seen in training, but that the neural network can still sometimes accurately predict, thanks to its character level modelling.}
    \label{tab:pie_accs}
\end{table}

\begin{table}[htbp]
    \centering  \small %
    \begin{tabular}{c||r|rrrrrrr}
     & \textbf{\textit{Marmot}} & 
     \multicolumn{6}{c}{\textbf{\textit{Pie}}} & \\ \hline
 & \textbf{base} & \textbf{base} & +aux & \textbf{wembs} & +aux & \textbf{bert} & +aux & \textit{support}\\ \hline \hline
All & 96.87 & 96.72 & 96.51 & 96.84 & \textbf{97.01} & 96.65 & 96.15 & 4181\\ 
Ambiguous tokens & NA & 91.86 & 91.43 & \textbf{92.40} & 92.29 & 91.76 & 90.36 & 934\\ 
Unknown tokens & 82.57 & \textbf{86.24} & \textbf{86.24} & 78.44 & 81.65 & 76.61 & 73.39 & 218
\end{tabular}
    
    \caption{Results obtained for POS by the \textit{Marmot} and \textit{Pie} models on the test set. For the \textit{Pie} models, we confronted trainings with or without morphology as auxiliary tasks.
    } 
    \label{tab:marmot-results}
\end{table}

\subsection{Out-of-domain tests}

To evaluate the ability of the best models to generalise for data from other periods, we conceived two out-of-domain corpora.
Since we want to evaluate generality with regard to diachronic, diaphasic and diagenic (i.e. gender based) variation, we have selected samples from 16th to the 20th century texts, either from theatre plays or from a variety of genres outside theatre, literary or practical (administrative texts, correspondence, etc.), from male and female authors, in order to have:
\begin{itemize}
    \item 20 samples of roughly 100 tokens for each century, 10 of theatre plays, 10 of a variety of other genres; 
    \item roughly as much tokens written by men or women for each century;
    \item a comparable distribution of tokens by genres for each century.
\end{itemize}
In addition, for the samples concerning 17th century theatre, we excluded verse comedies in general, and the authors from which were drawn our training corpus. 
A complete list is given in appendix~\ref{annexe:out_of_domain}. 

We evaluate the best models (\textbf{wembs} and \textbf{wembs+aux}) for lemma and for POS on the out-of-domain data (table~\ref{tab:outofdomain_lemma} and \ref{tab:outofdomain_pos}). 

\begin{table}[htbp]
    \centering \small %
    \begin{tabular}{c|rrrrr|r}
        \multicolumn{7}{c}{\textsc{Lemma}} \\ \hline \hline
        \textit{Corpus} & \textbf{16th} & \textbf{17th} & \textbf{18th} & \textbf{19th} & \textbf{20th} & \textit{All cent.} \\ \hline
        \textbf{Theatre} &  97.60&98.10&98.88&98.34&98.00&98.19\\
        \textbf{Not theatre} & 97.78&98.02&98.06&96.97&97.39&97.65\\
        \textbf{Both} & 97.69&98.06&98.46&97.66&97.70&97.92\\
    \end{tabular}
    \caption{Lemmatisation accuracies of the best models on out-of-domain data}
    \label{tab:outofdomain_lemma}
\end{table}

\begin{table}[htbp]
    \centering \small %
    \begin{tabular}{c|rrrrr|r}
        \multicolumn{7}{c}{\textsc{POS}} \\ \hline \hline
        \textit{Corpus} & \textbf{16th} & \textbf{17th} & \textbf{18th} & \textbf{19th} & \textbf{20th} & \textit{All cent.} \\ \hline
        \textbf{Theatre} & 95.05&96.59&95.98&94.81&93.57&95.18\\
        \textbf{Not theatre} &92.89&94.27&96.53&91.87&91.35&93.42\\
        \textbf{Both} & 93.93&95.44&96.27&93.36&92.48&94.30\\
    \end{tabular}
    \caption{POS accuracies of the best models on out-of-domain data}
    \label{tab:outofdomain_pos}
\end{table}

The lemmatisation model proves to be relatively robust: globally, the loss of accuracy is of roughly 1 percentage point, 
while it is closer to 3 percentage points for the POS model. This difference can be explained by the difference between the training corpora: the use of significant additional data to improve the efficiency of the lemmatisation model seems to reflect in its greater capacity to generalise. The same reason could also explain why the lemmatisation models transpose better to non-dramatic texts than the POS model.

In both cases, though, the better accuracies are reached for the 17th and 18th centuries --~and, surprisingly, more specifically for 18th century data in most cases. It progressively decreases going backward or forward in time.

\subsection{Improvement over previous corpus}

In order to evaluate the importance of the addition of the newly annotated data (\textit{Theatre}), we set-up a second training experiment where we reuse the best configuration found in \ref{sec:training-lemmatisation} (\textit{wembs}) over 5 iterations using the same training, dev and test set, except for the files from the \textit{Theatre} sample in train and dev. Using the best model, we are able to measure the effect of the introduction of \textit{Theatre} data on top of the \textit{Frantext} data. On lemmatization, the results are clear: the overall accuracy with the same configuration goes from 97.30\% to 99.09\%, an increase of 1.79 points. We cannot compare unknown and ambiguous tokens as well as unknown tokens as the support from these categories moved: -88 unknown tokens, +34 ambiguous tokens, -23 unknown targets. 
The same impact can be seen on out-of-domain data (Table \ref{tab:improvement}) with a more visible impact on the Theatre corpus of the 18th century (+0.96 points). Strangely enough, neither the 17th century or the Theatre genre benefits the most from the inclusion of 17th century Theatre data, as the \textit{Not Theatre} corpus and the 18th century are the most improved ones.

\begin{table}[htbp]
    \centering \small %
\begin{tabular}{l|rrrrr|r}
\textit{Corpus}               & \textbf{16} & \textbf{17} & \textbf{18} & \textbf{19} & \textbf{20} & \textit{All cent.} \\ \hline \hline
\textbf{Theatre}     & +0.16       & +0.30       & +0.96       & +0.00       & +0.22       & +0.32              \\
\textbf{Not theatre} & +0.74       & +0.46       & +0.63       & +0.30       & +0.67       & +0.56              \\
\textbf{Both}        & +0.46       & +0.38       & +0.79       & +0.07       & +0.44       & +0.45             
\end{tabular}
\caption{Score improvements between \textit{Frantext} corpus and \textit{Frantext}+\textit{Theatre}. Reads as follow: on 17th century theatre, a model trained with \textit{Frantext+Theatre} has +0.30 points of accuracy, going from 97.80 to 98.10\%.}
\label{tab:improvement}
\end{table}

\subsection{Most frequent confusions}

The most frequent confusions of the best models on the out-of-domain data are presented in table~\ref{tab:pie_confMat}. 

\begin{table}[htbp]
    \centering 
\begin{footnotesize}
\begin{tabular}[t]{cccc}
\textbf{Expected}&\textbf{Tot. Err.}&\textbf{Pred.}&\textbf{Pred. times}\\ \hline \hline
\multicolumn{4}{c}{\textsc{Lemma}}\\ \hline
 monsieur&9&M.&9 \\ 
 le&9&il&9 \\ 
 cœur&6&cuur&4 \\&&crur&2 \\ 
 franc&6&Franc&6 \\ 
 Électre&5&éLECTRE&4 \\&&électre&1 \\ 
 noble&4&Nobles&4 \\ 
 maître&4&maîtresse&4 \\ 
 de\_le&4&un&4 \\ 
 un&4&de\_le&4 \\ 
 dame&3&Dame&2 \\&&damer&1 \\ 
 Phanor&3&phanor&3 \\ 
 voir&3&vivre&3 \\ 
 Vosges&3&vosge&3 \\ 
 Médée&3&médé&3
 \end{tabular}
 \begin{tabular}[t]{cccc}
\textbf{Expected}&\textbf{Tot. Err.}&\textbf{Pred.}&\textbf{Pred. times}\\ \hline \hline
 \multicolumn{4}{c}{\textsc{POS}}\\ \hline
 NOMcom&182&ADJqua&45 \\ 
&&NOMpro&33 \\ 
&&VERinf&32 \\ 
&&VERcjg&18 \\ 
&&VERppe&11 \\ 
 ADJqua&104&NOMcom&45 \\ 
&&VERppe&17 \\ 
&&VERcjg&12 \\ 
 NOMpro&81&NOMcom&43 \\ 
 VERcjg&61&NOMcom&18 \\ 
&&ADJqua&12 \\ 
&&VERppe&12 \\ 
 ADVgen&58&NOMcom&16 \\ 
&&VERcjg&14 \\ 
 PROrel&31&CONsub&26
\end{tabular}
\end{footnotesize}
    \caption{Sample from the confusion matrix for the best \textit{Pie} models on the out-of-domain data. }
    \label{tab:pie_confMat}
\end{table}

Regarding lemmatisation, some errors related to homographs such as the token \textit{le} (regimen pronoun (\textit{il}) or determiner (\textit{le}); or the token \textit{des} (plural of the determiner \textit{un} or partitive \textit{de\_le}). Some other errors are due to abbreviations not present in the training data (\textit{M.} for \textit{monsieur}). More interestingly, for the research of potential improvements, many errors are related to rare characters classes in the training data, such as capital letters or ligatures (\textit{œ}).

For POS, the most frequent confusions are in nominal rather than verbal tags. In particular, confusions between common nouns (\texttt{NOMcom}), proper nouns (\texttt{NOMpro}), adjectives (\texttt{ADJqua}) and nominal forms of the verbs (participle, \texttt{VERppe}, and infinitive, \texttt{VERinf}). Some errors are due to the morpho-syntactic nature of our annotation, which, for instance, labels substantive adjective as common nouns (\textit{le \textbf{beau}} is \texttt{NOMcom}).

\section{Discussion}
\strut

The scores of up to 99\% for in-domain lemmatisation and 97\% for POS tagging put our best models above the expected state of the art, or in it's upper range (see \ref{sec:introduction}). Yet, these results were deserving of further investigation, through out-of-domain tests and more qualitative inspection.

Out-of-domain tests show that, even though they were trained for 17th century classical theatre, the model reach their best accuracies for 18th century texts. Such a fact seems counter-intuitive, but a plausible explanation might be as follow: the median date of the texts of the \textit{Frantext} corpus is 1872, and mean date is 1859. By adding roughly 80k tokens of 17th century texts to the 2,3M of Frantext (with mean text date 1859), we may have somehow slightly pulled the corpus closer to the 18th century.

In any case, through the careful construction of a labelled corpus and the use of a neural lemmatiser and tagger, we were able to develop models very suitable to the annotation necessary for the stylometric analyses showed in \cite{Cafieroeaax5489}. But we think that the model are useful for the annotation of 17th century theatre, and even beyond that, of Early Modern French texts in normalised spelling, with encouraging results regarding generalisation beyond the original scope of the experiments.

\section{Further research}
\strut

Looking at our results, the main lead for improvements should be a more efficient way to deal with rare character classes, such as capital letters, diacritics or ligatures. Several methods could be used to reduce the number of classes (\textit{e.g.}, Unicode decomposed normalisation) or, alternatively, the training set could be sufficiently extended to provide enough cases.

More generally, three possible directions could be followed in the coming years. The first is to expand the training corpus to other dramatic genres (tragedy, tragi-comedy\dots), and other genres in general (poetry, novels, short stories…). The second would be to replace normalised tokens by non-normalised ones, and therefore offer a new model that takes full advantage of the ability of tools like \textit{Pie} to deal with spelling variation in historical languages, and, doing so, strengthen the ability of model to deal with older varieties of French. 
The third is to expand dramatically the training with data taken from 18th or 16th\,c. texts. Results going in these directions and continuing the experiments described in this paper will be presented at DTUC'20~\citep{gabay_standardizing_2020}.

\section*{Author contributions}
\strut

P.F. encoded the corpus and all its metadata. F.C., J.-B. C. and S.G. designed the research project. The preprocessing of the texts, the initial setup of the \textit{Pyrrha} corpus and of the authority lists were performed by J.-B. C. Correction of the training data and expansion of the authority lists was shared equally between F.C., J.-B. C. and S.G. Post-processing of the trained corpus, injection of morphological data, and correction of the \textit{Frantext} data was done by J.-B. C., as well as the training, testing of the models and their further use to annotate unseen data. T.C. programmed modifications to \textit{Pie} code to include \textit{Bert} and participated in the training and benchmarking of models, as well as additional debugging of the annotation tools.
All authors contributed to the writing of this paper.

The authors have no competing interests to declare.

\section*{Materials and data availability}

\strut

The most up-to-date version of the models can be easily obtained and used thanks to the \texttt{pie-extended} Python package, available on Pypi (\url{https://pypi.org/project/pie-extended/}), with the command \texttt{pie-extended download fr}, and can be queried at \url{https://tal.chartes.psl.eu/deucalion/}.

The initial version created for \cite{Cafieroeaax5489} is available from the \textit{Science Advances} website since the publication of the paper, on 27th Nov. 2019 at this address: \href{https://advances.sciencemag.org/highwire/filestream/221312/field_highwire_adjunct_files/0/aax5489_Data_file_S1.zip}{\texttt{https://advan\-ces.science\-mag.org/high\-wire/file\-stream/221312/field\_highwire\_ad\-junct\_fi\-les/0/aax5489\_Data\_file\_S1.zip}}. The
initial version of the models is available on Zenodo (\href{http://dx.doi.org/10.5281/zenodo.3353421}{10.5281/zenodo.3353421}).

The version of the best models described in this paper, as well as training, validation and test data can be found on Zenodo as well, as version 3 of the same repository (doi: \href{http://dx.doi.org/10.5281/zenodo.3828644}{10.5281/zenodo.3828644}).%

Licensing is specified separately for each repository.

\section*{Acknowledgements}
\strut

We thank the DIM \textit{Science du texte et connaissances nouvelles} for funding the acquisition of a GPU server, as well as the École nationale des chartes for providing infrastructure and support for the server.
We thank Enrique Manjavacas for his precious advice regarding lemmatisation and Pie configuration, as well as Marie Puren for her help with the annotation of 20th century texts.

\bibliographystyle{plainnat}
\bibliography{corpusAndModels}

\begin{thebibliography}{50}
\providecommand{\natexlab}[1]{#1}
\providecommand{\url}[1]{\texttt{#1}}
\expandafter\ifx\csname urlstyle\endcsname\relax
  \providecommand{\doi}[1]{doi: #1}\else
  \providecommand{\doi}{doi: \begingroup \urlstyle{rm}\Url}\fi

\bibitem[Adda et~al.(1998)Adda, Mariani, Lecomte, Paroubek, and
  Rajman]{Adda1998}
Gilles Adda, Joseph Mariani, Josette Lecomte, Patrick Paroubek, and Martin
  Rajman.
\newblock The grace french part-of-speech tagging evaluation task.
\newblock In \emph{Proc. of LREC'98 (1st International Conference on Language
  Resources and Evaluation)}, 1998.
\newblock URL \url{https://infoscience.epfl.ch/record/98004}.

\bibitem[ATILF-CNRS and {Université de Lorraine}(1998-2018)]{frantext}
ATILF-CNRS and {Université de Lorraine}.
\newblock Base textuelle {F}rantext: Démonstration, 1998-2018.
\newblock URL \url{https://www.frantext.fr/repository/frantext-demo/}.

\bibitem[ATILF-CNRS and {Université de Lorraine}(2016)]{morphalou}
ATILF-CNRS and {Université de Lorraine}.
\newblock Morphalou v3.1, 2016.
\newblock URL \url{https://www.ortolang.fr/market/lexicons/morphalou}.

\bibitem[{Bibliothèque nationale de France}(1997)]{gallica}
{Bibliothèque nationale de France}.
\newblock Gallica - bibliothèque numérique de la bnf, 1997.
\newblock URL \url{https://gallica.bnf.fr}.

\bibitem[Bollmann(2019)]{bollmann_large-scale_2019}
Marcel Bollmann.
\newblock A {Large}-{Scale} {Comparison} of {Historical} {Text} {Normalization}
  {Systems}.
\newblock \emph{Proceedings of the 2019 Conference of the North}, pages
  3885--3898, 2019.
\newblock \doi{10.18653/v1/N19-1389}.
\newblock URL \url{http://arxiv.org/abs/1904.02036}.

\bibitem[Cafiero and Camps(2019)]{Cafieroeaax5489}
Florian Cafiero and Jean-Baptiste Camps.
\newblock Why {M}oli{\`e}re most likely did write his plays.
\newblock \emph{Science Advances}, 5\penalty0 (11), 2019.
\newblock \doi{10.1126/sciadv.aax5489}.
\newblock URL \url{https://advances.sciencemag.org/content/5/11/eaax5489}.

\bibitem[Camps(2019)]{camps_geste:_2019}
Jean-Baptiste Camps, editor.
\newblock \emph{Geste: un corpus de chansons de geste, 2016-… ({Version}
  02)}.
\newblock École nationale des chartes, Paris, April 2019.
\newblock URL \url{http://doi.org/10.5281/zenodo.2630574}.
\newblock textes du domaine public, développements CC-BY-SA.

\bibitem[Camps and Couffignal(2017)]{camps_production_2017}
Jean-Baptiste Camps and Gilles~Guilhem Couffignal.
\newblock La production de corpus d’occitan médiéval et prémoderne:
  problèmes et perspectives de travail.
\newblock In \emph{Actes du {XII}{\textbackslash}ieme\{\} {Congrès}
  {International} de l'{Association} {Internatioale} d'Études {Occitanes},
  {Albi} 2017}, 2017.
\newblock URL \url{https://halshs.archives-ouvertes.fr/halshs-02050089/}.

\bibitem[Cl\'{e}rice(2020)]{pie-extended}
Thibault Cl\'{e}rice.
\newblock pie-extended 0.0.13, 2020.
\newblock URL \url{https://pypi.org/project/pie-extended/}.

\bibitem[Cl\'{e}rice et~al.(2019)Cl\'{e}rice, Pilla, and
  Jean-Baptiste-Camps]{thibault_clerice_2019_2541730}
Thibault Cl\'{e}rice, Julien Pilla, and Jean-Baptiste-Camps.
\newblock hipster-philology/pyrrha: 2.0.0, 2019.
\newblock \url{https://doi.org/10.5281/zenodo.2541730}.

\bibitem[Clérice et~al.(2019)Clérice, Camps, and Pinche]{deucalionAF}
Thibault Clérice, Jean-Baptiste Camps, and Ariane Pinche.
\newblock Deucalion, modèle ancien francais, 2019.
\newblock URL \url{http://dx.doi.org/10.5281/zenodo.2539134}.

\bibitem[Dereza(2019)]{dereza2019lemmatisation}
Oksana Dereza.
\newblock Lemmatisation for under-resourced languages with sequence-to-sequence
  learning: A case of early irish.
\newblock In \emph{Proceedings of Third Workshop Computational linguistics and
  language science}, volume~4, pages 113--124, 2019.

\bibitem[Diwersy et~al.(2017)Diwersy, Falaise, Lay, and
  Souvay]{diwersy_ressources_2017}
Sascha Diwersy, Achille Falaise, Marie-Hélène Lay, and Gilles Souvay.
\newblock Ressources et méthodes pour l’analyse diachronique.
\newblock \emph{Langages}, N 206\penalty0 (2):\penalty0 21--44, August 2017.
\newblock ISSN 0458-726X.
\newblock URL \url{https://www.cairn.info/revue-langages-2017-2-page-21.htm}.

\bibitem[Duval(2015)]{duval_les_2015}
Frédéric Duval.
\newblock Les éditions de textes du {XVIIe} siècle.
\newblock In David Trotter, editor, \emph{Manuel de la philologie de
  l'édition}, pages 369--394. De Gruyter, 2015.
\newblock ISBN 978-3-11-030260-8.
\newblock \doi{10.1515/9783110302608-017}.
\newblock URL
  \url{https://www.degruyter.com/view/books/9783110302608/9783110302608-017/9783110302608-017.xml}.

\bibitem[Fièvre(2007)]{fievre_theatre_2007}
Paul Fièvre.
\newblock Théâtre classique, 2007.
\newblock URL \url{http://www.theatre-classique.fr}.

\bibitem[Gabay(2014)]{gabay_pourquoi_2014}
Simon Gabay.
\newblock Pourquoi moderniser l’orthographe? principes d’ecdotique et
  littérature du {XVIIe} siècle.
\newblock \emph{Vox Romanica}, 73\penalty0 (1):\penalty0 27--42, 2014.
\newblock ISSN 0042-899X.
\newblock URL
  \url{https://elibrary.narr.digital/article/99.125005/vox201410027}.

\bibitem[Gabay and Barrault(2020)]{gabay:hal-02596669}
Simon Gabay and Lo{\"i}c Barrault.
\newblock {Traduction automatique pour la normalisation du fran{\c c}ais du
  XVII e si{\`e}cle}.
\newblock In \emph{{TALN 2020}}, 27{\`e}me Conf{\'e}rence sur le Traitement
  Automatique des Langues Naturelles, Nancy, France, June 2020. {ATALA}.
\newblock URL \url{https://hal.archives-ouvertes.fr/hal-02596669}.

\bibitem[Gabay et~al.(2020{\natexlab{a}})Gabay, Camps, and Clérice]{gabay2020}
Simon Gabay, Jean-Baptiste Camps, and Thibault Clérice.
\newblock Manuel d'annotation linguistique pour le français moderne (xvie
  -xviiie siècles), 2020{\natexlab{a}}.
\newblock URL \url{https://hal.archives-ouvertes.fr/hal-02571190}.

\bibitem[Gabay et~al.(2020{\natexlab{b}})Gabay, Clérice, Camps, Tanguy, and
  Gille-Levenson]{gabay_standardizing_2020}
Simon Gabay, Thibault Clérice, Jean-Baptiste Camps, Jean-Baptiste Tanguy, and
  Matthias Gille-Levenson.
\newblock Standardizing linguistic data: method and tools for annotating
  (pre-orthographic) {French}.
\newblock In \emph{Proceedings of the 2nd {International} {Conference} on
  {Digital} {Tools} \& {Uses} {Congress}}, pages 1--7, 2020{\natexlab{b}}.

\bibitem[Guillot et~al.(2013{\natexlab{a}})Guillot, Prévost, and
  Lavrentiev]{guillot_manuel_2013}
Céline Guillot, Sophie Prévost, and Alexei Lavrentiev.
\newblock \emph{Manuel de référence du jeu {Cattex09}}.
\newblock École normale supérieure de Lyon, Lyon, 2013{\natexlab{a}}.
\newblock URL \url{http://bfm.ens-lyon.fr/IMG/pdf/Cattex2009_manuel_2.0.pdf}.
\newblock Version 2.0 – 8 avril 2013.

\bibitem[Guillot et~al.(2013{\natexlab{b}})Guillot, Prévost, and
  Lavrentiev]{prevost_principes_2013}
Céline Guillot, Sophie Prévost, and Alexei Lavrentiev.
\newblock Principes d’annotation cattex09.
\newblock Technical report, École normale supérieure de Lyon, Lyon,
  2013{\natexlab{b}}.
\newblock version 2.0.
  \url{http://bfm.ens-lyon.fr/IMG/pdf/Cattex2009_principes_2.0.pdf}.

\bibitem[Guillot-Barbance et~al.(2017)Guillot-Barbance, Heiden, and
  Lavrentiev]{Guillot2017}
Céline Guillot-Barbance, Serge Heiden, and Alexei Lavrentiev.
\newblock Base de français médiéval : une base de référence de sources
  médiévales ouverte et libre au service de la communauté scientifique.
\newblock \emph{Diachroniques}, 7:\penalty0 168--184, 2017.
\newblock URL \url{https://halshs.archives-ouvertes.fr/halshs-01809581}.

\bibitem[Heiden(2010)]{heiden_txm_2010}
Serge Heiden.
\newblock The {TXM} platform: {Building} open-source textual analysis software
  compatible with the {TEI} encoding scheme.
\newblock In \emph{24th {Pacific} {Asia} {C}onference on {L}anguage,
  {I}nformation and {C}omputation}, pages 389--398. DECODE, Waseda University,
  2010.

\bibitem[Heigold et~al.(2016)Heigold, Neumann, and van
  Genabith]{heigold_neural_2016}
Georg Heigold, Guenter Neumann, and Josef van Genabith.
\newblock Neural morphological tagging from characters for morphologically rich
  languages.
\newblock \emph{arXiv preprint arXiv:1606.06640}, 2016.

\bibitem[Ide and Veronis(1994)]{Ide1994}
Nancy Ide and Jean Veronis.
\newblock Multext: Multilingual text tools and corpora.
\newblock In \emph{COLING 1994 Volume 1: The 15th International Conference on
  Computational Linguistics}, 1994.
\newblock URL \url{https://www.aclweb.org/anthology/C94-1097}.

\bibitem[Kestemont and De~Gussem(2017)]{kestemont_integrated_2017}
Mike Kestemont and Jeroen De~Gussem.
\newblock Integrated {Sequence} {Tagging} for {Medieval} {Latin} {Using} {Deep}
  {Representation} {Learning}.
\newblock \emph{Journal of Data Mining \& Digital Humanities}, 2017.
\newblock URL \url{https://jdmdh.episciences.org/3835}.

\bibitem[Kestemont et~al.(2016)Kestemont, Pauw, Nie, and
  Daelemans]{kestemont_lemmatization_2016}
Mike Kestemont, Guy~de Pauw, Renske~van Nie, and Walter Daelemans.
\newblock Lemmatization for variation-rich languages using deep learning.
\newblock \emph{Digital Scholarship in the Humanities}, page fqw034, 2016.
\newblock ISSN 2055-7671, 2055-768X.
\newblock \doi{10.1093/llc/fqw034}.
\newblock URL
  \url{http://dsh.oxfordjournals.org/content/early/2016/08/26/llc.fqw034}.

\bibitem[Leech and Wilson(1996)]{Leech1996}
Geoffrey Leech and Andrew Wilson.
\newblock Eagles: Recommendations for the morphosyntactic annotation of
  corpora.
\newblock Technical report, Expert Advisory Group on Language Engineering
  Standards, 1996.
\newblock URL
  \url{https://home.uni-leipzig.de/burr/Verb/htm/LinkedDocuments/annotate.pdf}.

\bibitem[Ling et~al.(2015)Ling, Luís, Marujo, Astudillo, Amir, Dyer, Black,
  and Trancoso]{ling_finding_2015}
Wang Ling, Tiago Luís, Luís Marujo, Ramón~Fernandez Astudillo, Silvio Amir,
  Chris Dyer, Alan~W. Black, and Isabel Trancoso.
\newblock Finding function in form: {Compositional} character models for open
  vocabulary word representation.
\newblock \emph{arXiv preprint arXiv:1508.02096}, 2015.

\bibitem[{Manjavacas} et~al.(2019){Manjavacas}, {K{\'a}d{\'a}r}, and
  {Kestemont}]{pie}
Enrique {Manjavacas}, {\'A}kos {K{\'a}d{\'a}r}, and Mike {Kestemont}.
\newblock {Improving Lemmatization of Non-Standard Languages with Joint
  Learning}.
\newblock \emph{arXiv e-prints}, art. arXiv:1903.06939, Mar 2019.
\newblock URL \url{https://www.aclweb.org/anthology/N19-1153/}.

\bibitem[Manjavacas et~al.(2019)Manjavacas, Kádár, and
  Kestemont]{manjavacas_improving_2019}
Enrique Manjavacas, Akos Kádár, and Mike Kestemont.
\newblock Improving lemmatization of non-standard languages with joint
  learning.
\newblock \emph{arXiv preprint arXiv:1903.06939}, 2019.

\bibitem[{Martin} et~al.(2019){Martin}, {Muller}, {Ortiz Su{\'a}rez}, {Dupont},
  {Romary}, {Villemonte de la Clergerie}, {Seddah}, and {Sagot}]{camembert}
Louis {Martin}, Benjamin {Muller}, Pedro~Javier {Ortiz Su{\'a}rez}, Yoann
  {Dupont}, Laurent {Romary}, {\'E}ric {Villemonte de la Clergerie}, Djam{\'e}
  {Seddah}, and Beno{\^\i}t {Sagot}.
\newblock {CamemBERT: a Tasty French Language Model}.
\newblock \emph{arXiv e-prints}, art. arXiv:1911.03894, Nov 2019.

\bibitem[M{\"u}ller et~al.(2013)M{\"u}ller, Schmid, and
  Sch{\"u}tze]{muller2013efficient}
Thomas M{\"u}ller, Helmut Schmid, and Hinrich Sch{\"u}tze.
\newblock Efficient higher-order crfs for morphological tagging.
\newblock In \emph{Proceedings of the 2013 Conference on Empirical Methods in
  Natural Language Processing}, pages 322--332, 2013.

\bibitem[M{\"u}ller et~al.(2015)M{\"u}ller, Cotterell, Fraser, and
  Sch{\"u}tze]{muller-etal-2015-joint}
Thomas M{\"u}ller, Ryan Cotterell, Alexander Fraser, and Hinrich Sch{\"u}tze.
\newblock Joint lemmatization and morphological tagging with lemming.
\newblock In \emph{Proceedings of the 2015 Conference on Empirical Methods in
  Natural Language Processing}, pages 2268--2274, Lisbon, Portugal, September
  2015. Association for Computational Linguistics.
\newblock \doi{10.18653/v1/D15-1272}.
\newblock URL \url{https://www.aclweb.org/anthology/D15-1272}.

\bibitem[Petrov et~al.(2011)Petrov, Das, and McDonald]{Petrov2011}
Slav Petrov, Dipanjan Das, and Ryan~T. McDonald.
\newblock A universal part-of-speech tagset.
\newblock \emph{CoRR}, abs/1104.2086, 2011.
\newblock URL \url{http://arxiv.org/abs/1104.2086}.

\bibitem[Prevost Sophie, Stein Achim, SRCMF,
  2020()]{noauthor_universaldependenciesud_old_french-srcmf_2020}
Prevost Sophie, Stein Achim, SRCMF, 2020.
\newblock {UniversalDependencies}/{UD}\_old\_french-{SRCMF}, November 2020.
\newblock URL
  \url{https://github.com/UniversalDependencies/UD_Old_French-SRCMF}.
\newblock original-date: 2017-05-30T08:20:15Z.

\bibitem[Prévost and Stein(2013)]{prevost_syntactic_2013}
Sophie Prévost and Achim Stein, editors.
\newblock \emph{Syntactic {Reference} {Corpus} of {Medieval} {French}
  ({SRCMF})}.
\newblock ENS de Lyon; Lattice, Paris; ILR University of Stuttgart,
  Lyon/Stuttgart, v2.6 edition, 2013.
\newblock URL \url{http://srcmf.org}.

\bibitem[Prévost et~al.(2013)Prévost, Guillot, Lavrentiev, and
  Heiden]{prevost_jeu_2013}
Sophie Prévost, Céline Guillot, Alexei Lavrentiev, and Serge Heiden.
\newblock Jeu d’étiquettes morphosyntaxiques {CATTEX}2009.
\newblock Technical report, École normale supérieure de Lyon, Lyon, 2013.
\newblock version 2.0. \url{http://bfm.ens-lyon.fr/IMG/pdf/Cattex2009_2.0.pdf}.

\bibitem[Riffaud(2014)]{riffaud2014}
Alain Riffaud.
\newblock Répertoire du théâtre français imprimé au xviie siècle, 2014.
\newblock URL \url{https://repertoiretheatreimprime.yale.edu/}.

\bibitem[Romary et~al.(2004)Romary, Salmon-Alt, and Francopoulo]{Romary2004}
Laurent Romary, Susanne Salmon-Alt, and Gil Francopoulo.
\newblock Standards going concrete: from lmf to morphalou.
\newblock In \emph{The 20th International Conference on Computational
  Linguistics (COLING 2004) - ElectricDict '04 Proceedings of the Workshop on
  Enhancing and Using Electronic Dictionaries}, pages 22--28, 2004.
\newblock URL \url{https://hal.inria.fr/inria-00121489}.

\bibitem[Sagot(2016)]{sagot2016}
Beno{\^i}t Sagot.
\newblock {External Lexical Information for Multilingual Part-of-Speech
  Tagging}.
\newblock Research Report RR-8924, {Inria Paris}, June 2016.
\newblock URL \url{https://hal.inria.fr/hal-01330301}.

\bibitem[Sagot(2010)]{Sagot2010}
Benoît Sagot.
\newblock The le\textit{fff}, a freely available and large-coverage
  morphological and syntactic lexicon for french.
\newblock In \emph{Proceedings of the 7th international conference on Language
  Resources and Evaluation}, 2010.
\newblock URL \url{https://hal.archives-ouvertes.fr/inria-00521242}.

\bibitem[Schmid(1995)]{Schmid1995}
Helmut Schmid.
\newblock Improvements in part-of-speech tagging with an application to german.
\newblock In \emph{Proceedings of the ACL SIGDAT-Workshop}, page 47–50, 1995.

\bibitem[Schmid(2019)]{schmid2019deep}
Helmut Schmid.
\newblock Deep learning-based morphological taggers and lemmatizers for
  annotating historical texts.
\newblock In \emph{Proceedings of the 3rd International Conference on Digital
  Access to Textual Cultural Heritage}, pages 133--137, 2019.

\bibitem[Schöch(2018)]{schoch-2018}
Christof Schöch.
\newblock \textit{Théâtre Classique}, paul fièvre (ed.), 2007-2018.
\newblock \emph{RIDE}, 8, 2018.
\newblock URL \url{https://ride.i-d-e.de/issues/issue-8/theatre-classique}.

\bibitem[Souvay and Pierrel(2009)]{Souvay2009}
Gilles Souvay and Jean-Marie Pierrel.
\newblock Lgerm lemmatisation des mots en moyen français.
\newblock \emph{Traitement Automatique des Langues}, 50\penalty0 (2):\penalty0
  149--172, 2009.
\newblock URL \url{https://halshs.archives-ouvertes.fr/halshs-00396452}.

\bibitem[Tang et~al.(2018)Tang, Cap, Pettersson, and
  Nivre]{tang_evaluation_2018}
Gongbo Tang, Fabienne Cap, Eva Pettersson, and Joakim Nivre.
\newblock An {Evaluation} of {Neural} {Machine} {Translation} {Models} on
  {Historical} {Spelling} {Normalization}.
\newblock \emph{arXiv:1806.05210 [cs]}, 2018.
\newblock URL \url{http://arxiv.org/abs/1806.05210}.

\bibitem[Tellier et~al.(2012)Tellier, Dupont, and
  Courmet]{tellier2012segmenteur}
Isabelle Tellier, Yoann Dupont, and Arnaud Courmet.
\newblock Un segmenteur-{'e}tiqueteur et un chunker pour le fran{\c{c}}ais (a
  segmenter-pos labeller and a chunker for french)[in french].
\newblock In \emph{Proceedings of the Joint Conference JEP-TALN-RECITAL 2012,
  volume 5: Software Demonstrations}, pages 7--8, 2012.

\bibitem[Urieli(2013)]{urieli2013}
Assaf Urieli.
\newblock \emph{Robust French syntax analysis: reconciling statistical methods
  and linguistic knowledge in the Talismane toolkit}.
\newblock PhD thesis, Université Toulouse le Mirail - Toulouse II, 2013.
\newblock URL \url{https://tel.archives-ouvertes.fr/tel-00979681/}.

\bibitem[Vigier and Blumenthal(2013-2017)]{presto}
Denis Vigier and Peter Blumenthal.
\newblock Presto - l'évolution du système prépositionnel du français,
  2013-2017.
\newblock URL \url{http://presto.ens-lyon.fr}.

\end{thebibliography}

\appendix\footnotesize

\section{Plays sampled to create the initial training corpus}
\label{annexe:corpusMain} 

\vspace*{1cm}

The following plays selected from \cite{fievre_theatre_2007} were sampled.

\begin{tiny}\hspace*{-5cm}%
    		\begin{longtable}{lp{2cm}p{3cm}lp{1cm}lp{0.5cm}ll}
    			\textbf{id}&\textbf{auteur}&\textbf{titre}&\textbf{date}&\textbf{genre}&\textbf{inspiration}&\textbf{structure}&\textbf{type}&\textbf{periode}\\ \hline \hline
    			CP\_DONSANCHE&CORNEILLE, Pierre&DON SANCHE D'ARAGON, COMÉDIE HÉROÏQUE&1649&Comédie héroïque&moeurs espagnoles&5 act.&vers&1641-1650\\ 
    			CP\_GALERIEDUPALAIS&CORNEILLE, Pierre&LA GALERIE DU PALAIS ou L'AMIE RIVALE&1637&Comédie&moeurs françaises&5 act.&vers&1631-1640\\ 
    			CP\_ILLUSIONCOMIQUE&CORNEILLE, Pierre&L'ILLUSION COMIQUE, COMÉDIE&1639&Comédie&moeurs françaises&5 act.&vers&1631-1640\\ 
    			CP\_MELITE33&CORNEILLE, Pierre&MÉLITE OU LES FAUSSES LETTRES, COMÉDIE&1633&Comédie&moeurs françaises&5 act.&vers&1621-1630\\ 
    			CP\_MENTEUR&CORNEILLE, Pierre&LE MENTEUR, COMÉDIE&1644&Comédie&moeurs françaises&5 act.&vers&1641-1650\\ 
    			CP\_PULCHERIE&CORNEILLE, Pierre&PULCHÉRIE, COMÉDIE HÉROÏQUE&1673&Comédie héroïque&histoire chrétienne&5 act.&vers&1671-1680\\ 
    			CP\_SUITEMENTEUR&CORNEILLE, Pierre&LA SUITE DU MENTEUR, COMÉDIE&1645&Comédie&moeurs françaises&5 act.&vers&1641-1650\\ 
    			CP\_SUIVANTE&CORNEILLE, Pierre&LA SUIVANTE, COMÉDIE&1637&Comédie&moeurs françaises&5 act.&vers&1631-1640\\ 
    			CP\_TITE&CORNEILLE, Pierre&TITE ET BÉRÉNICE, COMÉDIE HEROÏQUE&1671&Comédie héroïque&histoire romaine&5 act.&vers&1661-1670\\
    			CP\_VEUVE34&CORNEILLE, Pierre&LA VEUVE OU LE TRAÎTRE TRAHI, COMÉDIE&1634&Comédie&moeurs françaises&5 act.&vers&1631-1640\\ 
    			CT\_AMOURALAMODE&CORNEILLE, Thomas&L'AMOUR À LA MODE, COMÉDIE. &1651&Comédie&moeurs espagnoles&5 act.&vers&1651-1660\\ 
    			CT\_CHARMEDELAVOIX&CORNEILLE, Thomas&LE CHARME DE LA VOIX, COMÉDIE&1658&Comédie&moeurs italiennes&5 act.&vers&1651-1660\\ 
    			CT\_COMTESSEORGUEIL&CORNEILLE, Thomas&LA COMTESSE D'ORGUEIL, COMÉDIE&1690\dag{}&Comédie&moeurs françaises&5 act.&vers&1651-1660\\ 
    			CT\_DOMBERTRANDECIGARRAL&CORNEILLE, Thomas&DON BERTRAN DE CIGARRAL, COMÉDIE&1709*&Comédie&moeurs espagnoles&5 act.&vers&1651-1660\\ 
    			CT\_DOMCESARDAVALOS&CORNEILLE, Thomas&DON CÉSAR D'AVALOS, COMÉDIE.&1661&Comédie&moeurs espagnoles&5 act.&vers&1671-1680\\ 
    			CT\_FEINTASTROLOGUE&CORNEILLE, Thomas&LE FEINT ASTROLOGUE, COMÉDIE&1651&Comédie&moeurs françaises&5 act.&vers&1651-1660\\ 
    			CT\_FESTINPIERRE&CORNEILLE, Thomas&LE FESTIN DE PIERRE, COMÉDIE&1677&Comédie&moeurs espagnoles&5 act.&vers&1671-1680\\ 
    			CT\_GALANTDOUBLE&CORNEILLE, Thomas&LE GALANT DOUBLÉ, COMÉDIE. &1659&Comédie&moeurs espagnoles&5 act.&vers&1651-1660\\ 
    			CT\_GEOLIERDESOISMEME&CORNEILLE, Thomas&LE GEÔLIER DE SOI-MÊME, COMÉDIE.&1655&Comédie&moeurs italiennes&5 act.&vers&1651-1660\\ 
    			CT\_ILLUSTRESENNEMIS&CORNEILLE, Thomas&LES ILLUSTRES ENNEMIS, COMÉDIE&1657&Comédie&moeurs espagnoles&5 act.&vers&1651-1660\\ 
    			CT\_INCONNU&CORNEILLE, Thomas&L'INCONNU, COMÉDIE.&1675&Comédie&moeurs françaises&5 act.&vers&1671-1680\\ 
    			M\_AMPHITRYON&MOLIÈRE&AMPHITRYON, COMÉDIE&1668&Comédie&mythe grec&3 act., prol.&vers&1661-1670\\ 
    			M\_DEPITAMOUREUX&MOLIÈRE&LE DÉPIT AMOUREUX&1656&Comédie&moeurs françaises&5 act.&vers&1651-1660\\ 
    			M\_DOMGARCIEDENAVARRE&MOLIERE&DON GARCIE DE NAVARRE, COMÉDIE&1682\ddag{}&Comédie&moeurs espagnoles&5 act.&vers&1661-1670\\
    			M\_ECOLEDESFEMMES&MOLIÈRE&L'ÉCOLE DES FEMMES, COMÉDIE.&1663&Comédie&moeurs françaises&5 act.&vers&1661-1670\\ 
    			M\_ETOURDI&MOLIÈRE&L'ÉTOURDI ou LES CONTRE-TEMPS, COMÉDIE&1663&Comédie&moeurs françaises&5 act.&vers&1661-1670\\
    			M\_FEMMESSAVANTES&MOLIÈRE&LES FEMMES SAVANTES, COMÉDIE&1672&Comédie&moeurs françaises&5 act.&vers&1671-1680\\ 
    			M\_MISANTHROPE&MOLIÈRE&LE MISANTHROPE ou L'ATRABILAIRE AMOUREUX, COMÉDIE&1667&Comédie&moeurs françaises&5 act.&vers&1661-1670\\ 
    			M\_TARTUFFE&MOLIÈRE&LE TARTUFFE ou L'IMPOSTEUR, COMÉDIE&1669&Comédie&moeurs françaises&5 act.&vers&1661-1670\\ 
O\_ABSENTCHEZSOI&OUVILLE, Antoine le Métel&L'ABSENT CHEZ SOI&1643&Comédie&moeurs françaises&5 act.&vers&1641-1650\\ 
O\_FAUSSESVERITES&OUVILLE, Antoine le Métel&LES FAUSSES VÉRITÉS, COMÉDIE&1643&Comédie&moeurs françaises&5 act.&vers&1641-1650\\ 
O\_SOUPCONS&OUVILLE, Antoine le Métel&LES SOUPÇONS SUR LES APPARENCES, COMÉDIE&1650&héroïco-comédie&moeurs françaises&5 act.&vers&1641-1650\\ 
    			R\_BELLEALPHREDE&ROTROU, Jean&LA BELLE ALPHRÈDE, COMÉDIE&1639&Comédie&moeurs arabes&5 act.&vers&1631-1640\\ 
    			R\_CAPTIFS&ROTROU, Jean&LES CAPTIFS OU LES ESCLAVES, COMÉDIE&1640&Comédie&moeurs françaises&5 act.&vers&1631-1640\\ 
    			R\_SOSIES&ROTROU, Jean&LES SOSIES, COMÉDIE&1638&Comédie&mythe grec&5 act.&vers&1631-1640\\ 
    			S\_DOMJAPHETDARMENIE&SCARRON, Paul&DON JAPHET D'ARMÉNIE, COMÉDIE.&1653&Comédie&moeurs françaises&5 act.&vers&1651-1660\\ 
    			S\_GARDIENDESOIMEME&SCARRON, Paul&LE GARDIEN DE SOI-MÊME&1654&Comédie&moeurs italiennes&5 act.&vers&1651-1660\\ 
    			S\_HERITIERRIDICULE&SCARRON, Paul&L'HÉRITIER RIDICULE OU LA DAME INTÉRESSÉE..&1650&Comédie&moeurs espagnoles&5 act.&vers&1641-1650\\ 
    			S\_JODELET&SCARRON, Paul&LE JODELET OU LE MAÎTRE VALET, COMÉDIE.&1648&Comédie&moeurs françaises&5 act.&vers&1641-1650\\ 
    			S\_JODELETDUELISTE&SCARRON, Paul&LE JODELET DUELLISTE, COMÉDIE.&1646&Comédie&moeurs françaises&5 act.&vers&1641-1650\\ 
    			S\_MARQUISRIDICULE&SCARRON, Paul&LE MARQUIS RIDICULE, OU LA COMTESSE faite à la hâte. COMÉDIE.&1656&Comédie&moeurs espagnoles&5 act.&vers&1651-1660
    		\end{longtable}
    	\end{tiny}

\dag{} The edition used as base is from 1690, but the play dates back to 1670.

\* The edition used as base is from 1709, but the play dates back to 1652.

\ddag{} Created in 1661, but published 1680.

\clearpage

    \section{Annotated texts from the \textit{Frantext} base}
    \label{annexe:frantext_openaccess}

\vspace*{1cm}

The following texts, from \cite{frantext}, were used as complementary data for lemmatisation training.

    	\begin{tiny} \hspace*{-2cm}%
    		\begin{longtable}{lllllllllll}
    			id&auteur&titre&date&place&publisher&N.tok\\ \hline \hline
    			K639:1890:52916&LOTI, Pierre&Le Roman d'un enfant&1891&Paris&Calmann-Levy&61,417\\
    			K934:1898:35797&ROSTAND Edmond&Cyrano de Bergerac&1898&Paris&Fasquelle&47,213\\
    			K999:1899:45606&GOURMONT Remy de&Esthétique de la langue française…
    			&1899&NA&Soc. Mercure de Fr.&53,640\\
    			L233:1895:13272&JARRY Alfred&Ubu Roi, \textit{Œuvres complètes}, t. 4.&s.d.&Monte-Carlo&éd. Du Livre&17,672\\
    			L266:1905:59455&POINCARÉ Henri&La Valeur de la science&1905&Paris&Flammarion&66,125\\
    			L433:1882:31745&BECQUE Henry&Les Corbeaux, \textit{Théâtre complet}, t. 2&1922&s.p.&Fasquelle&38.813\\
    			L486:1884:68155&HUYSMANS Joris-Karl&À rebours&1907&Paris&Fasquelle&79,058\\
    			L499:1901:63336&JAURÈS Jean&Études socialistes&1902&Paris&Ollendorf&71,159\\
    			L784:1908:100993&LEROUX Gaston&Le Parfum de la dame en noir&1908&Paris&L'Illustration&N.tok\\
    			L846:1857:68636&ABOUT Edmond&Le Roi des montagnes&1857&s.p.&Hachette&79,720\\
    			L884:1867:20871&MEILHAC Henri et Ludovic HALÉVY&La Vie parisienne&1867&Paris&M. Levy&27,381\\
    			M223:1873:70999&VERNE Jules&Le Tour du monde en quatre-vingts jours&s.d.&Paris&Hetzel&84,925\\
    			M289:1874:69484&FROMENTIN Eugène&Un été dans le Sahara&1877&Paris&Plon&80,612\\
    			M362:1825:111416&BRILLAT-SAVARIN Jean-Anthelme&Physiologie du goût… 
    	    	&1847&Paris&Charpentier&127,968\\
    			M374:1830:28471&FOURIER Charles&Le Nouveau monde industriel…
    			&1830&Paris&Bossange&32,566\\
    			M425:1794:4621&CHÉNIER André&L'Invention, \textit{Œuvres complètes}, t. 2.&1910&Paris&Delagrave&5,473\\
    			M433:1807:57291&STAËL Germaine de&Corinne ou l'Italie : t. 1&1807&s.p.&Peltier&65,012\\
    			M464:1801:72408&DESTUTT DE TRACY&Élémens d'idéologie, 1
    			&1804&Paris&Courcier&80,433\\
    			M468:1803:62746&KRÜDENER Barbara Juliane von&Valérie&1840&Paris&Charpentier&73,463\\
    			M473:1809:82847&LAMARCK Jean-Baptiste&Philosophie zoologique : t. 1&1809&s.p.&Dentu&95,111\\
    			M492:1805:123259&CUVIER Georges&Leçons d'anatomie comparée : t. 1&1805&Paris&Baudouin&138,674\\
    			M528:1798:12861&GUILBERT DE PIXERÉCOURT&Victor ou l'Enfant de la forêt&1798&Paris&Barba&16,878\\
    			M548:1832:199497&SAY Jean-Baptiste&Traité d'économie politique&1841&Paris&Guillaumin&223,038\\
    			M622:1789:24158&SIEYÈS Emmanuel&Qu'est-ce que le Tiers état ?&1888&Paris&Soc. Hist. Rév. Fr.&27,002\\
    			M629:1792:31251&FLORIAN Jean-Pierre&Fables&1792&Paris&Didot&37,172\\
    			M893:1869:61374&GONCOURT Edmond et Jules de&Madame Gervaisais&1876&s.p.&Charpentier&71,174\\
        		M939:1852:84555&COMTE Auguste&Catéchisme positiviste…
        		&1909&Paris&Garnier&95,589\\
    			N245:1838:32486&HUGO Victor&Ruy Blas, \textit{Œuvres complètes. Théâtre}, 3 &1905&Paris&Ollendorff&41,487\\
    		    N268:1802:63535&BONALD Louis de&Législation primitive… 
    		    t. 1&
    		    1802&Paris&Le Clère&72,539\\
    			N429:1778:64388&BUFFON Georges-Louis de&Des époques de la nature, \textit{Hist. Natur.}, t. 5.&1778&Paris&Impr. Royale&70,866\\
    			P556:1872:177870&VIOLLET-LE-DUC Eugène&Entretiens sur l'architecture : t. 2&1872&Paris&A. Morel&199,695\\
    			Q454:1784:53428&RÉTIF DE LA BRETONNE Nicolas&La Paysanne pervertie…
    			t. 1&1784&s.p.& Vve Duchesne&64,424\\
    		\end{longtable}
    	\end{tiny} 


\section{Out-of-domain texts}
\label{annexe:out_of_domain}

\vspace*{1cm}

Here follows the list of the texts that were sampled for building the out-of-domain test set. Spelling was modernised when necessary. 

Theatre texts were sampled from \cite{fievre_theatre_2007}, with the exception of:

\begin{description}
\item[AUBIGNAC\_PUCELLE] sample transcribed and modernised from the edition Paris, 1642, available on \textit{Gallica}.
\item[GRINGORE\_SAINT-LOUIS] sample transcribed  and modernised from the \textit{Œuvres complètes}, éd. Ch. d'Héricault et A. Montaiglon, Paris, 1858-1877, available on \textit{Gallica}.
\item[ANOUILH\_MEDEE] sample transcribed from the \textit{Nouvelles pièces noires}, Paris, La Table Ronde, 1976.
\item[GIRAUDOUX\_ELECTRE] sample transcribed from the \textit{Théâtre complet}, éd. J. Body, Paris, Gallimard, 1982.
\item[CESAIRE\_CHRISTOPHE] sample transcribed from \textit{Poésie, Théâtre, Essais et Discours}, éd. Albert James Arnold, Paris, CNRS Editions, 2013.
\end{description}

For the non-dramatic texts, the main sources, as shown in the table, are:
\begin{description}
\item[ELEC] \textit{Éditions en ligne de l'École des chartes}, \url{http://elec.enc.sorbonne.fr/}.
\item[GALL] \textit{Gallica}, Bibliothèque nationale de France, \url{https://gallica.bnf.fr}.
\item[WS] \textit{Wikisource: la bibliothèque libre}, \url{https://fr.wikisource.org/}.
\end{description}

Other online editions were occasionnaly used:

\begin{description}
\item[] \textit{Correspondance d'Isabelle de Charrière}, éd.
Suzan van Dijk and Madeleine van Strien-Chardonneau, 	\url{https://charriere.huygens.knaw.nl/}.
\item[] \textit{Les Nouvelles Nouvelles (1663) par Jean Donneau de Visé}, éd. Claude Bourqui et Christophe Schuwey, \url{http://www.unifr.ch/nouvellesnouvelles/}. 
\item[] 
\textit{Œuvres de Frédéric le Grand - Werke Friedrichs des Großen
Digitale Ausgabe der Universitätsbibliothek Trier}, dir. Hans-Ulrich Seifert, Trier, \url{http://friedrich.uni-trier.de/}.
\item[] \textit{Testaments de Poilus (1914-1918): transcription collaborative}, dir. Florence Clavaud, 2018-…, \url{https://testaments-de-poilus.huma-num.fr}.
\end{description}

The sample from MONTEGUT\_ISMENE was transcribed and modernised from the edition Paris, 1768, available on \textit{Google Books}.

\subsection{Theatre}

\hspace*{-5cm} \begin{tiny} %
\begin{longtable}{p{2.5cm}p{2cm}p{3cm}lp{1cm}lp{0.5cm}ll}
\textbf{id}&\textbf{author}&\textbf{title}&\textbf{date}&\textbf{genre}&\textbf{inspiration}&\textbf{structure}&\textbf{type}&\textbf{period}\\ \hline \hline
ANONYME\_PARDONNEUR&anonyme&FARCE NOUVELLE TRÈS BONNE ET FORT JOYEUSE À TROIS PERSONNAGES, FARCE&1500 c.&Farce&Sermon joyeux&1 act.&vers&1501-1600\\ 
ANONYME\_PONTAUXANES&anonyme&FARCE NOUVELLE FORT JOYEUSE DU PONT AUX ÂNES&1500 c.&Farce&Sermon joyeux&1 act.&vers&1501-1600\\ 
ANONYME\_RESURRECTION\-JENINLANDORE&anonyme&FARCE NOUVELLE TRÈS BONNE ET FORT JOYEUSE DE LA RESURRECTION DE JENIN LANDORE, FARCE&1500 c.&Farce&Sermon joyeux&1 act.&vers&1501-1600\\ 
ANONYME\_SERMONJOYEUX&anonyme&SERMON JOYEUX DE BIEN BOIRE, FARCE&1500 c.&Farce&Sermon joyeux&1 act.&vers&1501-1600\\ 
BEZE\_ABRAHAM&BEZE, Théodore&ABRAHAM SACRIFIANT, TRAGÉDIE.&1550&Tragédie&bible&1 act.&vers&1541-1550\\ 
GRINGORE\_SAINT-LOUIS&GRINGORE, Pierre&Mystère de Saint Louis&1514&mystère&histoire médiévale&8 livres&vers&1511-1520\\ 
JODELLE\_CLEOPATRE&JODELLE, Étienne&CLÉOPÂTRE CAPTIVE, TRAGÉDIE.&1574&Tragédie&histoire romaine&5 act.&vers&1571-1580\\ 
JODELLE\_DIDON&JODELLE, Étienne&DIDON SE SACRIFIANT, TRAGÉDIE.&1574&Tragédie&mythe grec&5 act.&vers&1571-1580\\ 
JODELLE\_EUGENE&JODELLE, Étienne&L'EUGÈNE, COMÉDIE.&1574&Comédie&moeurs françaises&5 act.&vers&1571-1580\\ 
TURNEBE\_CONTENTS&TURNÈBE, Odet de&LES CONTENTS, COMÉDIE NOUVELLE EN PROSE FRANÇAISE&1584&Comédie&moeurs françaises&5 act.&prose&1581-1590\\ \hline
AUBIGNAC\_PUCELLE&AUBIGNAC, François Hédelin&La Pucelle d'Orléans&1642&Tragédie&histoire médiévale&&prose&1641-1650\\ 
CHAMPREPUS\_ULYSSE&CHAMPREPUS, Jacques de&ULYSSE, TRAGÉDIE.&1603&Tragédie&mythe grec&5 act.&vers&1601-1610\\ 
CHAPPUZEAU\_ARMETZAR&Samuel Chappuzeau (1625-1701) &ARMETZAR OU LES AMIS ENNEMIS, TRAGICOMÉDIE.&1656&Tragi-comédie&histoire turque&5 act.&vers&1651-1660\\ 
DESHOULIERES\_GENSERIC&DESHOULIÈRES, Antoinette du Ligier de la Garde&GENSERIC, TRAGÉDIE&1680&Tragédie&histoire romaine&5 act.&vers&1671-1680\\ 
DESJARDINS\_FAVORI&DESJARDINS, Marie-Catherine-Hortense dite de Villedieu&LE FAVORI, TRAGICOMÉDIE.&1665&Tragi-comédie&moeurs espagnoles&5 act.&vers&1661-1670\\ 
DURANT\_OISIVETE&DURANT, Catherine&OISIVITÉ EST MÈRE DE TOUS LES VICES, PROVERBE.&1699&Proverbe&moeurs françaises&1 act.&prose&1691-1700\\ 
MATHIEU\_MAGICIENNE\-ETRANGERE&MATTHIEU, Pierre&LA MAGICIENNE ÉTRANGÈRE, TRAGÉDIE.&1617&Tragédie&histoire française&4 act.&vers&1611-1620\\ 
SCUDERY\_LIGDAMON\-ELIDIAS&SCUDERY, Georges de&LIGDAMON ET LIDIAS, TRAGI-COMÉDIE&1631&Tragi-comédie&histoire médiévale&5 act.&vers&1631-1640\\ 
URFE\_SYLVANIRE&URFÉ, Honoré d'&SYLVANIRE ou la MORTE VIVE, FABLE BOCAGÈRE&1627&Pastorale héroïque&pastorale&5 act., un prologue&vers&1621-1630\\ 
VONDREBECK-ALARD\_FORCESDELAMOUR&VONDREBECK, Maurice et ALARD, Charles&LES FORCES DE L'AMOUR ET DE LA MAGIE, DIVERTISSEMENT&1678&Divertissement&moeurs françaises&3 act.&prose&1671-1680\\ \hline
BERGASSE\_JOURNEE\-DESDUPES\_1790&BENSERADE, Isaac de&LA JOURNÉE DES DUPES, TRAGI-COMÉDIE&1790&Tragi-comédie&histoire française&4 act.&prose&1781-1790\\ 
BIEVRE\_VERCINGENTORIXE&BIÈVRE, François-Georges Mareschal de&VERCINGENTORIXE, TRAGÉDIE.&1770&Tragédie&histoire française&1 act.&vers&1761-1770\\ 
BOISSY\_VIEESTUNSONGE&BOISSY, Louis de&LA VIE EST UN SONGE, COMÉDIE-HÉROÏQUE.&1732&Comédie héroïque&moeurs polonaises&5 act.&vers&1731-1740\\ 
DANCOURT\_SANCHO&DANCOURT, Florent CARTON dit&SANCHO PANÇA, GOUVERNEUR, COMÉDIE&1712&Comédie&moeurs françaises&5 act.&vers&1711-1720\\ 
DIDEROT\_FILSNATUREL&DIDEROT, Denis&LE FILS NATUREL ou Les ÉPREUVES DE LA VERTU.&1757&Tragi-comédie&moeurs françaises&5 act.&prose&1751-1760\\ 
DUFRESNY\_MARIAGEFAIT\-ETROMPU&DUFRESNY, Charles&LE MARIAGE FAIT ET ROMPU, COMÉDIE&1721&Comédie&moeurs françaises&3 act.&vers&1721-1730\\ 
GUDIN\_LOTHAIRE&GUDIN de la BRENELLERIE, Paul-Philippe&LOTHAIRE, ROI DE LORRAINE, TRAGÉDIE&1759&Tragédie&histoire française&5 act.&vers&1751-1760\\ 
MARIVAUX\_PEREPRUDENT&MARIVAUX, Pierre de&LE PÈRE PRUDENT ET ÉQUITABLE, COMÉDIE&1712&Comédie&moeurs françaises&1 act.&vers&1711-1720\\ 
REGNARD\_MENECHMES&REGNARD, Jean-François&LES MÉNECHMES, ou LES JUMEAUX, COMÉDIE&1705&Comédie&moeurs françaises&5 act.&vers&1701-1710\\ 
VOLTAIRE\_MORTDECESAR&VOLTAIRE&LA MORT DE CÉSAR, TRAGÉDIE EN TROIS ACTES&1736&Tragédie&histoire romaine&3 act.&vers&1731-1740\\ \hline
ALLAIS\_BONBOUGRE&ALLAIS, Alphonse&LE PAUVRE BOUGRE ET LE BON GÉNIE, FÉÉRIE EN UN ACTE.&1889&Féérie&moeurs françaises&1 act.&prose&1881-1890\\ 
AUDE\_ECOLETRAGIQUE&AUDE, Joseph &L'ÉCOLE TRAGIQUE, OU CADET ROUSSEL MAITRE DE DÉCLAMATION COMÉDIE&1802&Comédie&moeurs françaises&1 act.&mixte&1801-1810\\ 
BANVILLE\_ANCIENPIERROT&BANVILLE, Théodore de&ANCIEN PIERROT, MONOLOGUE.&1877&Monologue&moeurs françaises&1 act.&prose&1871-1880\\ 
BONNETAIN\_APRESLE\-DIVORCE&BONNETAIN, Paul&APRÈS LE DIVORCE, PIÈCE&1890&Comédie&moeurs françaises&1 act.&prose&1881-1890\\ 
CONSTANT\_WALLSTEIN&CONSTANT, Benjamin&WALLSTEIN, TRAGÉDIE&1809&Tragédie&histoire allemande&5 act.&vers&1801-1810\\ 
DUMAS\_DONJUAN&DUMAS, Alexandre&DON JUAN DE MARANA, MYSTÈRE&1836&Drame&moeurs espagnoles&5 act.&mixte&1831-1840\\ 
GENLIS\_BELLEETLABETE&GENLIS, Stéphanie-Félicité Du Crest de&LA BELLE ET LA BÊTE, COMÉDIE&1829&Comédie&moeurs françaises&2 act.&prose&1821-1830\\ 
HUGO\_HERNANI&HUGO, Victor&HERNANI, OU L'HONNEUR CASTILLAN&1830&Drame&histoire espagnole&5 act.&vers&1821-1830\\ 
JARRY\_UBUROI&JARRY, Alfred&UBU ROI, DRAME&1896&Drame&Fantaisie historique&5 act.&mixte&1881-1890\\ 
SAND\_MOLIERE&SAND, George&MOLIÈRE, DRAME.&1851&Comédie&histoire littéraire&2 act.&prose&1851-1860\\ \hline
ANOUILH\_MEDEE&ANOUILH, Jean&Médée&1946&tragédie&mythe grec&NA&prose&1941-1950\\ 
BERNARDT\_PARTIEDE\-BRIDGE&BERNARD, Tristan&LA PARTIE DE BRIDGE, COMÉDIE&1930&Comédie&moeurs françaises&1 act.&prose&1921-1930\\ 
BERTON\_HOMMEQUIA\-TUELAMORT&BERTON, René&L'HOMME QUI A TUÉ LA MORT, DRAME&1928&Pièce dramatique&moeurs françaises&2 act.&prose&1921-1930\\ 
CESAIRE\_CHRISTOPHE&CÉSAIRE, Aimé&La Tragédie du roi Christophe&1963&tragédie&histoire haïtienne&3 act.&prose&1961-1970\\ 
COURTELINE\_MONSIEUR\-BADIN&COURTELINE, Georges&MONSIEUR BADIN, COMÉDIE.&1904&Saynète&moeurs françaises&1 act.&prose&1891-1900\\ 
GIRAUDOUX\_ELECTRE&GIRAUDOUX, Jean&Électre&1937&tragédie bourgeoise&mythe grec&2 act.&prose&1931-1940\\ 
HUGUES\_CENDRILLON&HUGUES, Clovis&CENDRILLON, COMÉDIE.&1906&Comédie&conte de fées&1 act.&vers&1901-1910\\ 
HUGUES\_TYL&HUGUES, Clovis&TYL L'ESPIEGLE, COMÉDIE.&1906&Comédie&moeurs flamande&1 act.&vers&1901-1910\\ 
LESENNE\_REVEIL\-CORNEILLE&LE SENNE, Camille&LE RÉVEIL DE CORNEILLE, POÈME DRAMATIQUE.&1916&Dialogue des morts&histoire littéraire&1 act.&vers&1911-1920\\ 
RENARD\_MAITRESSE&RENARD, Jules&LA MAÎTRESSE, COMÉDIE.&1927&Comédie&moeurs françaises&2 act.&prose&1921-1930
\end{longtable}
 \end{tiny}

\subsection{Varia}

\begin{tiny} \hspace*{-5cm}%
\begin{longtable}{p{2.5cm}p{2cm}p{3cm}llllp{3cm}}
\textbf{ID}&\textbf{auteur}&\textbf{titre}&\textbf{genre}&\textbf{forme}&\textbf{date}&\textbf{source}&\textbf{ed.}\\ \hline \hline
ANONYME\_PAIX-BERGERAC&Chancellerie royale&Paix de Bergerac. Édit de Poitiers. Poitiers, septembre 1577.&légal&prose&1577&ELEC&éd. Bernard Barbiche, École nationale des chartes.\\ 
ANONYME\_REITRES&anonyme&Le vray discours sur la route et admirable desconfiture des Reistres…&discours&prose&1587&WS&éd. Éd. Fournier, Var. hist. et litt., t. 9\\ 
ARBEAU\_BELLE&Thoinot Arbeau&Belle qui tiens ma vie&chanson&vers&1588&WS&manq.\\ 
FRANCOISPREMIER\-\_CORRESP&François Ier&La correspondance du chancelier Antoine Du Bourg (1535-1538)&correspondance&prose&1537&ELEC&éd. Olivier Poncet\\ 
LENONCOURT-MARTIN\_HYPNEROTO\-MACHIE&Robert de Lenoncourt, Jean Martin (trads)&Hypnérotomachie, ou Discours du songe de Poliphile…&discours&prose&1546&WS&Jacques Kerver, 1546\\ 
LIEBAUT\_MISERES&Liébaut,Nicole Estienne&Les Misères de la Femme mariée…&poésie morale&vers&1587&WS&éd. Éd. Fournier, Var. hist. et litt., 3, 1855.\\ 
LOUISELABE\_ELEGIE&Louise Labé&Élégie I&poésie lyrique&vers&1555&WS&éd. Charles Boy, 1887.\\ 
MARGUERITENAVAR\-RE\_MIROIR&Marguerite de Navarre&Miroir de l'âme pécheresse&miroir&vers&1529&WS&éd. Félix Frank, 1873\\ 
RABELAIS\_PANTAGRUEL&Rabelais&Pantagruel&roman&prose&1532&WS&éd. Marty-Laveaux, 1868\\ 
RONSARD\_SONNETS&Ronsard, Pierre de&Je plante en ta faveur cet arbre de Cybèle&poésie&vers&1578&WS&éd. Roger Sorg,1921\\ \hline
ANNEROHAN\_PORTRAIT&Anne de Rohan-Soubise&Portrait de feue la duchesse de Nevers…&poésie&vers&1629&GALL&éd. Éd. de Barthélemy, Paris, 1862\\ 
ANONYME\_AIRABOIRE&anonyme&Air à boire&chanson&vers&16…&WS&La Chanson française du XVe au XXe siècle, éd. Jean Gillequin, 1910.\\ 
ANONYME\_ARREST&anonyme&Arrest notable donné au profit des femmes contre l’impuissance des maris&légal&prose&1626&WS&éd. Éd. Fournier, Var. hist. et litt., 6, 1856\\ 
BOILEAU\_SATIRE&Nicolas Boileau&satires (satire I)&satire&vers&1660&WS&Paris, 1872\\ 
DONNEAU\_NOUVELLES&Jean Donneau de Visé&Nouvelles nouvelles&actualité&prose&1663&other&éd. Bourqui \& Schuwey\\
ELISABETHBOHEME\-\_CORRESPONDANCE&Élisabeth de Bohême&Correspondance avec René Descartes&correspondance&prose&1643&WS&manq.\\ 
PERRAULT\_PETIT-CHAPERON&Charles Perrault&Le petit Chaperon rouge&conte&prose&1697&WS&Paris, 1902\\ 
POULLAIN\_EGALITE&François Poullain de La Barre&De l'Égalité des deux sexes, Discours phisique et moral&discours&prose&1679&GALL&Paris, 1679\\ 
SCUDERY\_ARTAMENE&Madeleine de Scudéry&Artamène ou le Grand Cyrus&roman&prose&1654&WS&Auguste Courbé, 1654.\\ 
VERVILLE\_UN-JOUR&Béroalde de Verville&« Un jour reconnaissant que je suis incapable »&poésie&vers&av. 1626&WS&manq.\\ \hline
ANONYME\_DECLARATION&anonyme&Déclaration des Droits de l’Homme et du Citoyen de 1793&légal&prose&1793&WS&manq.\\ 
CHARRIERE\_LETTRE&Isabelle de Charrière&lettre&correspondance&prose&1755&other&éd. S. van Dijk et M. van Strien-Chardonneau\\ 
CHATELET\_DISCOURS&Émilie du Châtelet&Discours sur le bonheur&discours&prose&1744&WS&éd. Bourlet de Vauxcelles, 1796\\ 
CHENIER\_ODE&André Chénier&Ode IX (à Charlotte Corday)&poésie&vers&1793&WS&éd. H. de Latouche, 1819\\ 
FREDERICDEUX\_DIVINE\-EMILIE&Frédéric II de Prusse&À la Divine Émilie&poésie&vers&1737&FG&Œuvres de Frédéric le Grand, Berlin, 1850\\ 
MERICOURT\_DISCOURS&Méricourt, Théroigne de&Discours devant la Société fraternelle des Minimes, 25 mars 1792&discours&prose&1792&GALL&Paris, 1792\\ 
MONTEGUT\_ISMENE&Jeanne de Montégut-Ségla&Ismene, élégie&poésie&vers&1739&other&Paris, 1768\\ 
ROUGET\_MARSEILLAISE&Rouget de Lisle&La Marseillaise&chanson&vers&1792&WS&manq.\\ 
TENCIN\_SIEGE&Madame de Tencin&Le Siège de Calais&roman&prose&1739&WS&éd. L.-S. Auger, Paris, 1820.\\ 
VOLTAIRE\_ZADIG&Voltaire&Zadig&nouvelle&prose&1747&WS&Paris, 1877.\\ 
\hline
CMH\_PV&Commission des monuments historiques&PV 29 février 1884&administratif&prose&1884&ELEC&éd. J.-D. Pariset.\\ 
DELESCLUZE\_LETTRE&Henri Delescluze&Lettre à Charles Delescluse (Carnets de prison)&correspondance&prose&1851&ELEC&éd. Chr. Nougaret et Fl.Clavaud\\ 
DEROULEDE\_CHANTS&Paul Déroulède&Nouveaux chants du soldat&chanson&vers&1883&GALL&Paris, 1883\\ 
DESBORDESVALMORE\_IN\-SOMNIE&Marceline Desbordes-Valmore&L'Insomnie&poésie&vers&1860&WS&Paris, 1860.\\ 
HEREDIA\_EPEE&Hérédia, José Maria de&L'Épée&poésie&vers&1893&WS&Paris, 1893.\\ 
MERIMEE\_ACADEMIE&Prosper Mérimée&Discours de réception à l’Académie française&discours&prose&1845&other&Site de l’Académie française\\ 
POTTIER\_INTERNATIONALE&Eugène Pottier&L'Internationale&chanson&vers&1871&WS&Paris, 1908.\\ 
SAND\_MARIANNE&George Sand&Marianne&roman&prose&1877&WS&Paris, 1877\\ 
SEGUR\_BOSSU&Comtesse de Ségur&François le Bossu&roman&prose&1864&WS&Paris, 1901\\ 
STAEL\_ALLEMAGNE&Mme de Staël&De l’Allemagne&essai&prose&1810&WS&Paris, 1814\\ 
\hline
ANONYME\_CONSCRITS&anonyme&Les Conscrits insoumis&chanson&vers&1902&WS&manq.\\ 
BLUM\_FRONTPOP&Léon Blum&« Nous sommes un gouvernement de front populaire »&discours&prose&1936&GALL&J. O.de la Rép. fr. 7 juin 1936\\ 
BRASILLACH\_JEANNE&Robert Brasillach&Le Procès de Jeanne d’Arc&étude historique&prose&1941&GALL&Paris, 1941\\ 
BRIMONT\_MIRAGES&Renée de Brimont&Mirages&poésie&vers&1919&WS&Paris, 1919\\ 
COLETTE\_MAISON&Colette&La Maison de Claudine&nouvelle&prose&1922&WS&manq.\\ 
LAHIRE\_ROUE&Jean de La Hire&La Roue fulgurante&feuilleton SF&prose&1908&GALL&Le Matin, 11 avril 1908\\ 
LASTEYRIE\_TESTAMENT&Gaspard Louis Guy de Lasteyrie marquis du Saillant&Testament&légal&prose&1915&other&Testaments de Poilus\\ 
LONDRES\_BAGNE&Albert Londres&Au bagne&reportage&prose&1924&WS&Paris, 1924\\ 
NOAILLES\_COEUR&Anna de Noailles&Le Coeur innombrable&poésie&vers&1901&WS&Paris, 1901\\ 
RACHILDE\_GRENOUILLES&Rachilde&Le Tueur de Grenouilles&nouvelle&prose&1900&WS&Mercure de France
\end{longtable}
\end{tiny}

\end{document}